%% file: clean.tex
\documentclass[11pt]{article}
\usepackage[T1]{fontenc}
\usepackage[utf8]{inputenc}
\usepackage{setspace}
\usepackage{amsfonts}
\usepackage{amsthm}
\usepackage{amsmath}
\usepackage{enumitem}
\usepackage{afterpage}
\usepackage{mathtools}
\usepackage{lscape}
\usepackage{subcaption}
\DeclareMathAlphabet{\mathdutchcal}{U}{dutchcal}{m}{n}
\usepackage{algorithm}
\usepackage{algorithmic}
\DeclareCaptionSubType*{algorithm}

\DeclareCaptionLabelFormat{alglabel}{Algorithm~#2}

\usepackage{float}
\usepackage[table]{xcolor}
\usepackage[margin=1in]{geometry}
\usepackage{soul}
\usepackage[normalem]{ulem}
\usepackage{cancel}
\usepackage{authblk}
\usepackage{xcolor}
\usepackage{natbib}
\usepackage{longtable}
\usepackage{array}
\usepackage{graphicx}

\newcommand{\z}{\mathbf{z}}

\newcommand{\C}{\mathbf{C}}

\newcommand{\D}{\mathbf{D}}
\newcommand{\A}{\mathbf{A}}

\newcommand{\B}{\mathbf{B}}
\newcommand{\G}{\mathbf{G}}

\newcommand{\bTheta}{\boldsymbol{\Theta}}
\newcommand{\bz}{\mathbf{z}}
\newcommand\bell{\boldsymbol{\ell}}
\newcommand\bxi{\boldsymbol{\xi}}

\newcommand{\Omegab}{\boldsymbol{\Omega}}
\newcommand{\bOmega}{\boldsymbol{\Omega}}

\newcommand{\omjj}[1]{\omega_{#1 #1}}
\newcommand{\omdotj}[1]{\boldsymbol{\omega}_{\bullet{#1}}}

\newcommand{\thetaj}[1]{\boldsymbol{\theta}_{#1}}

\newcommand{\tthetadotj}[1]{\Tilde{\boldsymbol{\theta}}_{\bullet{#1}}}

\newcommand{\I}{\mathbf{I}}

\newcommand{\norm}[1]{\left\| #1 \right\|}

\newcommand{\Vol}{\mathrm{Vol}}

\newtheorem{lemma}{Lemma}
\newtheorem{proposition}{Proposition}

\newtheorem{remark}{Remark}

\newtheorem{corollary}{Corollary}

\newcommand{\diag}{\mathrm{diag}}

\newcommand{\bL}{\mathbf{L}}
\newcommand{\bM}{\mathbf{M}}
\newcommand{\q}{\mathbf{q}}

\usepackage{hyperref}
\definecolor{darkblue}{rgb}{0.0,0.0,0.7}
\hypersetup{colorlinks,breaklinks,linkcolor=darkblue,urlcolor=darkblue,
            anchorcolor=darkblue,citecolor=darkblue}
            
\input{math_commands_JMLR}

\renewcommand\det{\mathrm{det}}

\numberwithin{theorem}{section}
\numberwithin{proposition}{section}
\numberwithin{lemma}{section}
\numberwithin{corollary}{section}

\title{The Reverse Telescoping Coordinate System for Positive Definite Matrices: Geometry, Computation, and Generative Modeling}
\author{Anindya Bhadra\footnote{Correspondence: bhadra@purdue.edu}}
\affil{Department of Statistics, Purdue University}
\date{}

\begin{document}

\maketitle

\begin{abstract}
We design a new unconstrained coordinate system where a $p\times p$ symmetric positive definite (SPD) matrix $\boldsymbol\Theta$ is represented by a \emph{reverse telescoping} map $\boldsymbol\Theta(\mathbf{x})=\mathcal{RT}(\mathbf{x})$, with $\mathbf{x}=(v,\mathbf{d},\mathbf{r})\in\mathbb{R} \times \mathbb{R}^{(p-1)}\times \mathbb{R}^{p(p-1)/2}$, representing respectively the log volume or log determinant; and the shape, as encoded by log relative diagonal scales and partial covariances among the nodes. This construction results in important properties not available in other charts, e.g., matrix logarithm, such as Jacobian depending on only the log-determinant. A useful feature of our construction is $\bx$ contains a \emph{lossless symbolic representation} of both the matrix and its inverse. Many important computations involving a matrix and its inverse can be performed in $\mathcal{O}(p^2)$ in the transformed domain, while it is the \emph{rendering of results in matrix forms} (on demand) that must incur an $\mathcal{O}(p^3)$ cost. Moreover, two unit-determinant matrices in the transformed domain can be joined by a straight line with pathwise unit determinant. For generative modeling, this allows designing a split volume--shape flow model trained by conditional flow matching for transporting the shape over the unit-determinant path, with a separate one-dimensional flow for transporting the volume or the determinant. The forbidding SPD \emph{constraint}, tamed thus into a powerful guiding force, leads to the surprising insight that it is in some sense \emph{easier} to design a volume-normalized shape flow for SPD compared to the unconstrained $\mathbb{R}^{p\times p}$, with no intrinsic notion of \emph{volume} to aid normalization, unlike the determinant of SPD matrices. We apply our construction for up to $p=200$ in generative modeling of SPD matrices on a difficult synthetic bimodal target, and in generating brain connectivity networks by models trained on fMRI data; as well as in intrinsic diffusion on the SPD manifold.
\end{abstract}

\section{Introduction}
Symmetric positive definite (SPD) matrix-valued data often arise from \emph{connectivity analysis}, e.g., in diffusion tensor imaging \citep[see, e.g.,][]{tournier2011diffusion,lope2018diffusion} or in the study of brain functional connectivity networks \citep[see, e.g.,][]{smith2013resting,smith2013functional,wei2024analyzing}, among others, where the \emph{covariance or partial covariance structure} is the primary object of interest, or is indeed the primary data, rather than the raw node-wise signals. Similarly, multivariate financial market data often display correlation patterns among component stocks \citep{mantegna1999hierarchical, laloux1999noise}. While model-based analysis of such data, e.g., learning a population covariance or precision matrix given some training samples using a parametric or nonparametric model is more classical \citep[see, e.g.,][]{pourahmadi2011covariance}, a key recent focus area of increasing importance is \emph{generative modeling} of such manifold-valued data \citep[see][for a recent survey]{loaiza-ganem2024deep}. That is, given only some samples from some underlying unknown SPD-valued data distribution, could one generate data with similar features, but not identical to, the training set? Given that SPD matrices form a Riemannian manifold \citep[][Section 1.1.2.E]{amari2016information}, a principled analysis of such data, either in a model-based or a generative framework, typically requires an unconstrained coordinate system or \emph{chart} that respects the SPD geometry. Among these, some of the most popular ones are Cholesky, matrix log and spectral or eigenvalue charts \citep{pennec2020manifold}. The overall contributions of this work stem from the construction of a new coordinate system based on recursive Schur updates to remove the SPD constraint, that we term the \emph{reverse telescoping} (RT) coordinates. The single-most distinguishing feature of this new chart is the isolation of the Jacobian into an intrinsic scalar coordinate encoding the log determinant of an SPD matrix, that, to our knowledge, does not occur under other commonly used charts. This leads to several nontrivial consequences, both geometric and algorithmic. Among the most important geometric consequences are a stage-local structure of the Fisher--Rao metric \citep{rao1945information} in the RT coordinates, a block diagonal inverse metric tensor, and that the Fisher--Rao volume form is a constant multiple of the Euclidean volume in normalized volume--shape RT coordinates. Among the most prominent algorithmic consequences are a separation of volume (encoded by the determinant) and shape (encoded by partial covariances and relative diagonal scales) of a matrix for learning a conditional flow matching model that allows these models to scale up to 200-dimensional matrices, while geometric obstacles become challenging for alternative methods using other charts at more than 50 dimensions. Similarly, the RT coordinates allow defining intrinsic diffusion on the SPD manifold with the same ease as the Euclidean space, with no additional bookkeeping cost for repeated evaluation of the metric tensor or chart Jacobian. Given the breadth of the current work, we provide the requisite background on the geometric and algorithmic aspects of generative SPD-modeling in the rest of this section, before summarizing our key contributions.

\subsection{The SPD Manifold and Existing Coordinate Charts}
A main challenge with modeling manifold-valued data, compared to real-valued data, lies in having to respect the manifold constraint or geometry. For the SPD manifold, various  coordinate systems or charts exist that allow unconstrained modeling of such data, while admitting a diffeomorphism between $\mathcal{M}_p^{+}$, the space of $p\times p$ SPD matrices, and $\mathbb{R}^{p(p+1)/2}$, the space of reals with the same dimension as the number of free elements in the matrix. For example, for $\bTheta\in\mathcal{M}_p^{+}$, the Cholesky decomposition allows one to write $\bTheta=\bL\bL^\top$, where $\bL$ is lower triangular with positive diagonal entries. Thus, for $i,j=1,\ldots,p;\; i\le j,$ if one defines $\bz=\{l_{ij}, \log l_{jj}\} \in \mathbb{R}^{p(p+1)/2}$, then one obtains a diffeomorphism under the log-Cholesky chart between $\mathcal{M}_p^{+}$ and $\mathbb{R}^{p(p+1)/2}$ and unconstrained modeling of real-valued $\bz$ is possible \citep{pourahmadi1999joint, lin2019}. Other charts that also allow unconstrained modeling include the matrix logarithm and log-spectral charts.

However, a key challenge under any chart is to relate its properties with the SPD space under the pullback. Several properties depend on the Jacobian of transformation, and this is where the difficulties begin to appear. For example, under the Cholesky chart mentioned above, one has:
\begin{align}
J (\mathbf{L}\mapsto\bTheta)= 2^p \prod_{j=1}^p l_{jj}^{p-j+1},\label{eq:jchol}
\end{align}
using standard calculations. One may immediately notice the unequal weighting of the diagonal $l_{ii}$ terms. Thus, the change that is induced for a perturbation in $l_{11}$ is vastly different from a perturbation in $l_{pp}$, i.e., the Jacobian is highly anisotropic and not compressible into a single native scalar chart coordinate. Similar issues arise with other existing charts. Table~\ref{tab:standard-chart-jacobians} gives a summary of the commonly used charts and their respective Jacobians. Here $\bTheta\in\mathcal M_p^+$, constants depending only on $p$ are suppressed,
and all Jacobians are with respect to Lebesgue measure on the symmetric matrix
entries. The formulas show that, standard charts either
weight coordinates anisotropically or couple all eigenvalues globally. Consequently, in standard charts the determinant, shape, and reference volume are globally entangled at the coordinate level, making likelihood evaluation, volume-normalized transport, and intrinsic volume preservation dependent on nontrivial Jacobian bookkeeping rather than a single native coordinate correction.
\begin{table}[t]
\centering

\footnotesize
\begin{tabular}{lll}
\toprule
Chart & Mapping & Chart Jacobian \\
\midrule
Cholesky &
$\bTheta = \L\L^\top,\quad l_{jj}>0$ &
$J(\L\mapsto\bTheta)
=
2^p\prod_{j=1}^p l_{jj}^{\,p-j+1}$ \\[0.8em]

Log-Cholesky &
$\bTheta=\L\L^\top,\quad l_{jj}=\exp(a_j)$ &
$J((a,\L_{\mathrm{off}})\mapsto\bTheta)
=
2^p\exp\!\left\{\sum_{j=1}^p (p-j+2)a_j\right\}$ \\[0.8em]

Modified Cholesky &
$\bTheta = \L\D\L^\top,\quad d_{jj}>0$ &
$J((\L,\D)\mapsto\bTheta)
=
\prod_{j=1}^p d_{jj}^{p-j}$ \\[0.8em]

Matrix logarithm &
$\bTheta=\exp(\A),\quad \A=\A^\top$ &
$J(\A\mapsto\bTheta)
=
\exp\{\operatorname{tr}(\A)\}
\prod_{i<j}
\frac{\exp(\alpha_i)-\exp(\alpha_j)}
{\alpha_i-\alpha_j}$ \\[1.0em]

Spectral  &
$\bTheta=\Q\bLambda \Q^\top,\quad \bLambda=\operatorname{diag}(\lambda_1,\ldots,\lambda_p)$ &
$J((\Q,\bLambda)\mapsto\bTheta)
\propto
\prod_{i<j}|\lambda_i-\lambda_j|$ \\[0.8em]

Log-spectral &
$\bTheta=\Q\operatorname{diag}(\exp(\alpha_1),\ldots,\exp(\alpha_p))\Q^\top$ &
$J((\Q,\alpha)\mapsto\bTheta)
\propto
\exp\!\left\{\sum_{i=1}^p \alpha_i\right\}
\prod_{i<j}|\exp(\alpha_i)-\exp(\alpha_j)|$ \\[1.0em]
\bottomrule
\end{tabular}
\caption{Common coordinate maps for SPD matrices and their chart Jacobians. Here $\alpha_1,\ldots,\alpha_p$ are the eigenvalues of $\A$ in the matrix-log
row, %and $\mu_1,\ldots,\mu_p$ are the eigenvalues of $B$ in the square-root row. 
and in the Jacobian formula, the ratio is interpreted by continuity when
$\alpha_i=\alpha_j$. Spectral coordinates are only locally one-to-one after
fixing an eigenvalue ordering and eigenvector sign convention.\label{tab:standard-chart-jacobians}
}
\end{table}
\normalsize

\subsection{Existing Generative Approaches for SPD-valued Data}
Generative modeling is a fast-paced world, and any cursory survey of influential methods is bound to be incomplete. Nevertheless, among the most influential approaches historically, the field had moved from early approaches based on Boltzmann machines or their variants \citep{ackley1985learning, salakhutdinov2009deep}, to variational autoencoders or VAEs \citep{kingma2013auto}, generative adversarial networks or GANs \citep{goodfellow2014generative}, score-based diffusion models \citep{song2020score} and flow-matching approaches \citep{lipman2022flow}; among many others. 

However, the outlook is more sobering under manifold-valued data  \citep{loaiza-ganem2024deep}, which can be attributed to geometric constraints. For general Riemannian manifolds, score matching \citep{debortoli2022riemannian} or flow matching \citep{chen2024flow} approaches exist, that essentially replace the Euclidean loss by a Riemannian metric, but their scalability is nowhere close to unconstrained modeling. 

Methods specialized to SPD manifolds have also appeared in the recent generative literature. \citet{li2024sdp} design a score-based denoising diffusion probabilistic model (DDPM) for SPD-valued data. \citet{collas2025riemannian} show that a conditional flow matching (CFM) under an appropriate pullback can successfully be used for generative modeling for SPD-valued data. \citet{tan25} use manifold-aware Wasserstein GANs for generating SPD-valued brain functional connectivity data. Scalability under geometric constraints remains a barrier for all methods.

\subsection{Summary of Our Contributions and Connections to Prior Works}
Our main contribution, interpreted at its broadest, is a new unconstrained coordinate system that we term the \emph{reverse telescoping coordinates}, admitting a diffeomorphism between $\mathcal{M}_p^{+}$ and $\mathbb{R}^{p(p+1)/2}$. The proposed coordinate system has its origin in the telescoping block decomposition of \citet{bhadra24jmlr} that was developed for computing evidence of marginal likelihood under Gaussian graphical models, and was used by \citet{gao2025order} for designing faster posterior sampling algorithms for such models. Some of the properties we present, such as unit Jacobian under a suitable parameterization, were noted by \citet{gao2025order}. However, these works  are solely concerned with posterior sampling of Bayesian graphical models, and neither work explored the reverse telescoping decomposition in its full generality to design an unconstrained coordinate system for SPD matrices itself. Our specific contributions could be summarized as follows.
\subsubsection{Geometric Contributions}
\begin{enumerate}
\item We demonstrate in Section~\ref{sec:rt} that in addition to being unconstrained, the RT coordinates admit an extremely simple form of the Jacobian, depending only on the determinant of an SPD matrix. Moreover, under a suitable reparameterization, the Jacobian is unity. Thus, all tasks involving a change of variables and Jacobian bookkeeping, e.g., evaluating or tracking the likelihood of SPD-valued data, become far simpler compared to all other commonly used charts. Specifically, for a map between $\mathbf{x}=(v,\mathbf{d},\mathbf{r})\in\mathbb{R} \times \mathbb{R}^{(p-1)}\times \mathbb{R}^{p(p-1)/2}$ and  $\bTheta\in\mathcal{M}_p^{+}$, we show that:
\begin{align}
J(\bx\mapsto\bTheta)=\exp(v),\label{eq:jrt}
\end{align}
where the scalar $v$ is the log-determinant of $\bTheta$. Contrasting~\eqref{eq:jrt} with~\eqref{eq:jchol} is revealing, as it shows the Cholesky anisotropy problem can be effectively resolved by a coordinate change. Comparisons with other charts in Table~\ref{tab:standard-chart-jacobians} lead to the same insight.

\item A further useful feature is that the same encoded object $\bx$ decodes losslessly to both $\bTheta$ and $\bTheta^{-1}$, so that likelihoods, quadratic forms, conditional quantities, and inverse-space computations can often be performed without repeatedly rendering or inverting dense matrices.

\item A direct consequence is the blockwise partitioning of the Fisher--Rao metric by freezing the past that does not introduce a global coupling. We contrast this explicitly with Fisher--Rao metric Cholesky coordinates in Section~\ref{sec:fish}. Another critical consequence is that under a suitable reparameterization, the Fisher--Rao volume density in normalized  RT coordinates becomes a constant multiple of the ambient Euclidean volume density, so that there is no volume distortion.

\item The Laplace--Beltrami operator and the intrinsic coordinates are explicitly derived in Section~\ref{sec:lb} that make designing an intrinsic diffusion as simple as a Euclidean diffusion.

\end{enumerate}

\subsubsection{Computational Contributions and Applications to Generative Modeling}
\begin{enumerate}

\item A critical property of the RT coordinates is the separation of volume and shape, that allows us to connect any two unit-determinant SPD matrices in the RT space by a straight line with pathwise unit-determinant. In Section~\ref{sec:app1}, we utilize this  property to design a split volume--shape flow model that can be trained by conditional flow matching. The volume--shape separation allows us to scale our approach to $p=200$ in generative SPD models, and an application to brain imaging data is described in Section~\ref{sec:brain}.

\item Similarly, as a geometric consequence of the identified intrinsic coordinates, we also design an intrinsic Langevin diffusion model as described in Lemma~\ref{prop:diff} for a given target law, which is useful in Bayesian inverse problems for sampling from a target posterior, and is a counterpart to the generative application described above.

\item For a $p\times p$ matrix, the encoding and decoding steps cost $\mathcal{O}(p^3)$. However, we show that once encoded, many commonly used operations involving SPD matrices, e.g., recurring computation of quadratic forms needed for density evaluation, can be performed in $\mathcal{O}(p^2)$ complexity. In particular, under spectral control, we obtain in Section~\ref{sec:app2} explicit upper and lower bounds on the Fisher--Rao distance computable at $\mathcal{O}(p^2)$, whereas exact computation remains $\mathcal{O}(p^3)$.

\end{enumerate}

\section{The Reverse Telescoping Coordinate System}\label{sec:rt}
\subsection{Notations}
 We use uppercase-bold for matrices, lowercase-bold for vectors and lowercase-unbolded for scalars. We denote by $\mathbf{A}_{j \times j}$ the upper-left $(j\times j)$ sub-matrix of a $p\times p$ matrix $\mathbf{A}_{p \times p}$ and $\mathbf{a}_j=(\mathbf{a}_{\bullet j}, a_{jj}) \in\mathbb{R}^{j-1}\times \mathbb{R}$ denotes the last column of $\mathbf{A}_{j \times j}$. When needed, $\mathbf{A}_{j:p \times j:p}$ denotes the lower-right $(p-j+1)\times (p-j+1)$ block of $\A$. We drop the subscripts by simply writing $\mathbf{A}$ when there is no scope for confusion, and otherwise make it explicit.

\subsection{The Encoding and Decoding Maps}
We parameterize an SPD matrix $\boldsymbol\Theta\in\mathcal M_{p}^{+}$ via an invertible map $\boldsymbol\Theta(\bx)=\mathcal{RT}(\mathbf{x})$, for $\mathbf{x}\in\mathcal{X}$,with $ \mathbf{x}=(v,\mathbf{d},\mathbf{r})$ a vector of length $p(p+1)/2$, equal to the number of free entries in $\boldsymbol\Theta$, where:
\begin{enumerate}
\item $v\in\mathbb{R}$, a scalar \emph{volume} coordinate equal to $\log \det(\bTheta)$,
\item $\mathbf{d}=(d_2,\ldots,d_{p})^{\top}\in\mathbb{R}^{(p-1)}$, a vector encoding \emph{log relative diagonal scales},
\item $\mathbf{r}= (\mathbf{r}^{\top}_{2},\ldots, \mathbf{r}^{\top}_{p})^{\top}\in\mathbb{R}^{p(p-1)/2}$, a vector encoding \emph{partial covariances}, where $\mathbf{r}_j$ is a vector of length $j-1$.
\end{enumerate}
To describe the encoding and decoding maps $\mathbf{x}\leftrightarrow\boldsymbol\Theta, \boldsymbol\Theta^{-1}$ we first introduce the intermediate object $\boldsymbol{\tilde\Theta}$. The encoding $\{\mathcal{T}:\boldsymbol\Theta\mapsto \boldsymbol{\tilde\Theta}\}$ and decoding $\{\mathcal{RT}: \boldsymbol{\tilde\Theta} \mapsto \boldsymbol\Theta, \boldsymbol\Theta^{-1}\}$ are based on the following invertible maps, as described in Algorithms~\ref{tab:fwd} and~\ref{tab:inv}. The encoding map, based on \emph{recursive} rank-one Schur complements,  is identical for the two algorithms, but is nevertheless repeated for the ease of reading side-by-side with the decoders. Although superficially this may look similar to rank-one Cholesky downdates, our use of Schur complements, and not Cholesky, plays a critical distinguishing role in the subsequent developments in this paper (elaborated upon later in Remark~\ref{rem:rtchol}). We note that Algorithm~\ref{tab:fwd} was introduced by \citet{bhadra24jmlr} and \citet{gao2025order} in order to facilitate sampling from the posterior under a Gaussian graphical model, but was not explored as a stepping stone to a generic unconstrained coordinate system for SPD matrices.

\setcounter{table}{0}

\begin{table}[!h]%
\small
\hspace{0.8cm}
\begin{subalgorithm}{.5\textwidth}
\begin{algorithmic}[1]
 \REQUIRE  $\boldsymbol{\Theta}_{p \times p}$, an SPD matrix with upper\\ triangle $(\boldsymbol\theta_1, \ldots,\boldsymbol\theta_p)$.
\STATE Initialize $\boldsymbol{\Omega}_{p \times p} \leftarrow \boldsymbol{\Theta}_{p \times p}$.
\STATE Set $\tilde{\boldsymbol{\theta}}_{p}=({\boldsymbol{\omega}}_{\bullet p}, {\omega}_{pp})^{\top}$.
\FOR {($j=p-1,\ldots, 1$)}
\STATE Update $\bOmega_{j\times j} \leftarrow \bOmega_{j\times j} - \frac{\tilde{\theta}_{\bullet(j+1)}\,\tilde{\btheta}_{\bullet(j+1)}^{\top}}{\tilde{\btheta}_{j+1,j+1}}$. 
\STATE Set $\tilde{\boldsymbol{\theta}}_{j}=(\omdotj{j}, \omjj{j})^{\top}$.
\ENDFOR
\ENSURE $\tilde{\boldsymbol{\Theta}}_{p \times p}$, a symmetric matrix with upper \\triangle $(\tilde{\boldsymbol\theta}_1, \ldots,\tilde{\boldsymbol\theta}_p)$.
\end{algorithmic}
\caption{Encoding:  $\boldsymbol\Theta_{p\times p}\mapsto\boldsymbol{\tilde\Theta}_{p\times p}$.}\label{algo_fwd}
\end{subalgorithm}%
%\hspace{2mm}
\begin{subalgorithm}{.5\textwidth}
\begin{algorithmic}[1]
\REQUIRE $\tilde{\boldsymbol{\Theta}}_{p \times p}$, a symmetric matrix with upper \\triangle $(\tilde{\boldsymbol\theta}_1, \ldots,\tilde{\boldsymbol\theta}_p)$.
\STATE Initialize $\tilde{\boldsymbol{\Omega}}_{p \times p} \leftarrow \tilde{\boldsymbol{\Theta}}_{p \times p}$.
\STATE Set ${\boldsymbol{\theta}}_{p}=(\tilde{\boldsymbol{\omega}}_{\bullet p}, \tilde{\omega}_{pp})^{\top}$.
\FOR {($j=p-1,\ldots, 1$)}
\STATE Update $\tilde{\bOmega}_{j\times j} \leftarrow \tilde{\bOmega}_{j\times j} + \frac{\tilde{\btheta}_{\bullet(j+1)}\,\tilde{\btheta}_{\bullet(j+1)}^{\top}}{\tilde{\theta}_{j+1,j+1}}$. 
\STATE Set ${\boldsymbol{\theta}}_{j}=(\tilde{\boldsymbol{\omega}}_{\bullet j}, \tilde{\omega}_{jj})^{\top}$.
\ENDFOR
 \ENSURE $\boldsymbol{\Theta}_{p \times p}$, an SPD matrix with upper \\triangle $(\boldsymbol\theta_1, \ldots,\boldsymbol\theta_p)$. 
\end{algorithmic}
\caption{Decoding:  $\boldsymbol{\tilde\Theta}_{p\times p}\mapsto \boldsymbol\Theta_{p\times p}$.}\label{algo_inv}
\end{subalgorithm}%
\captionsetup{labelformat=alglabel}
\caption{Encoder and decoder for the original matrix $\boldsymbol\Theta.$\\}%
\label{tab:fwd}%
\vspace{-0.1cm}
\end{table}

\begin{table}[!h]%
\small
\hspace{0.8cm}
\begin{subalgorithm}{.5\textwidth}
\begin{algorithmic}[1]
 \REQUIRE  $\boldsymbol{\Theta}_{p \times p}$, an SPD matrix with upper\\ triangle $(\boldsymbol\theta_1, \ldots,\boldsymbol\theta_p)$.
\STATE Initialize $\boldsymbol{\Omega}_{p \times p} \leftarrow \boldsymbol{\Theta}_{p \times p}$.
\STATE Set $\tilde{\boldsymbol{\theta}}_{p}=({\boldsymbol{\omega}}_{\bullet p}, {\omega}_{pp})^{\top}$.
\FOR {($j=p-1,\ldots, 1$)}
\STATE Update $\bOmega_{j\times j} \leftarrow \bOmega_{j\times j} - \frac{\tilde{\btheta}_{\bullet(j+1)}\,\tilde{\btheta}_{\bullet(j+1)}^{\top}}{\tilde{\theta}_{j+1,j+1}}$. 
\STATE Set $\tilde{\boldsymbol{\theta}}_{j}=(\omdotj{j}, \omjj{j})^{\top}$.
\ENDFOR
\ENSURE $\tilde{\boldsymbol{\Theta}}_{p \times p}$, a symmetric matrix with upper \\triangle $(\tilde{\boldsymbol\theta}_1, \ldots,\tilde{\boldsymbol\theta}_p)$.
\end{algorithmic}
\caption{Encoding:  $\boldsymbol\Theta_{p\times p}\mapsto\boldsymbol{\tilde\Theta}_{p\times p}$ (same as Algorithm~\ref{tab:fwd}(a)).}\label{algo_fwd}
\end{subalgorithm}%
%\hspace{2mm}
\begin{subalgorithm}{.5\textwidth}
\begin{algorithmic}[1]
\REQUIRE $\tilde{\boldsymbol{\Theta}}_{p \times p}$, a symmetric matrix with upper \\triangle $(\tilde{\boldsymbol\theta}_1, \ldots,\tilde{\boldsymbol\theta}_p)$.
\STATE Initialize ${\boldsymbol{\Sigma}}_{p \times p} \leftarrow \mathbf{0}_{p\times p}.$
\STATE Set $\sigma_{11}={\boldsymbol{\Sigma}}_{1 \times 1} = 1/\tilde{{\theta}}_{11}$.
\FOR {($j=1,\ldots, p-1$)}
\STATE Set $\bbeta_{\bullet(j+1)}=\frac{\tilde\btheta_{\bullet(j+1)}}{\tilde\theta_{j+1,j+1}}.$
\STATE Set $\bsigma_{\bullet(j+1)}=-{\boldsymbol{\Sigma}}_{j \times j}\bbeta_{\bullet(j+1)}$.
\STATE Set ${{\sigma}}_{j+1,j+1}= \frac{1}{\tilde\theta_{j+1,j+1}} + \bbeta^{\top}_{\bullet(j+1)} {\boldsymbol{\Sigma}}_{j \times j}\bbeta_{\bullet(j+1)}$.
\STATE Update ${\boldsymbol{\Sigma}}_{(j+1) \times (j+1)} \leftarrow \begin{bmatrix}
    {\boldsymbol{\Sigma}}_{j \times j} & \bsigma_{\bullet(j+1)}\\
    \bsigma_{\bullet(j+1)}^{\top} & {\sigma}_{j+1,j+1}
\end{bmatrix}.
$
\ENDFOR
 \ENSURE $\boldsymbol{\Sigma}_{p \times p}=\boldsymbol{\Theta}^{-1}_{p \times p}$, an SPD matrix with upper triangle $(\boldsymbol\sigma_1, \ldots,\boldsymbol\sigma_p)$. 
\end{algorithmic}
\caption{Decoding:  $\boldsymbol{\tilde\Theta}_{p\times p}\mapsto \boldsymbol\Theta_{p\times p}^{-1}$.}\label{algo_inv}
\end{subalgorithm}%
\captionsetup{labelformat=alglabel}
\caption{Encoder for $\boldsymbol\Theta$ and decoder for $\boldsymbol\Theta^{-1}$.\\}%
\label{tab:inv}%
%\vspace{0.2cm}
\end{table}
It is evident that all encoding and decoding algorithms are $\mathcal{O}(p^3)$. To see the correctness of the decoding steps, note from the two respective decoders in Algorithms~\ref{tab:fwd}(b) and~\ref{tab:inv}(b) that we have for $j=2,\ldots,p$:
\begin{equation}
\boldsymbol\Theta_{j} \leftarrow \begin{bmatrix} \boldsymbol\Theta_{j-1} + \frac{\tilde{\btheta}_{\bullet j}\,\tilde{\btheta}_{\bullet j}^{\top}}{\tilde{\theta}_{jj}} & \tilde{\btheta}_{\bullet j} \\ \tilde{\btheta}_{\bullet j}^\top & \tilde{\theta}_{jj} \end{bmatrix};\; \;
\boldsymbol\Sigma_{j} = \boldsymbol\Theta_{j}^{-1} \leftarrow \begin{bmatrix} \boldsymbol\Sigma_{j-1} & -\boldsymbol\Sigma_{j-1} \frac{\tilde\btheta_{\bullet j}}{\tilde\theta_{jj}} \\ -\frac{{\tilde\btheta_{\bullet j}}^\top}{\tilde{\theta}_{jj}} \boldsymbol\Sigma_{j-1} & \frac{1}{\tilde{\theta}_{jj}} + \frac{\tilde\btheta_{\bullet j}^\top \bSigma_{j-1} \tilde\btheta_{\bullet j}}{\tilde{\theta}_{jj}^2} \end{bmatrix},\label{eq:recursion}
\end{equation}
where we have abbreviated $\boldsymbol\Theta_{j}=\boldsymbol\Theta_{j\times j}$ and $\boldsymbol\Sigma_{j}=\boldsymbol\Sigma_{j\times j}$. That these two matrices are inverse of each other can be verified either via direct calculations or via an application of Equation (12) of \citet{guttman1946enlargement}. Suppose $p=1$. Then, by construction, the decoding algorithm sets $\theta_{11}=1/\sigma_{11}=\tilde\theta_{11}$. The correctness for an arbitrary positive integer $p$ now follows from an induction argument using the $p=1$ case and the recursion of~\eqref{eq:recursion}. The properties of Schur complements ensure positive definiteness for all $j$. This is because of the equivalence:
$$
\{\boldsymbol\Theta_{j} \in \mathcal{M}_{j}^{+}\}\Longleftrightarrow\{\boldsymbol\Theta_{j-1} \in \mathcal{M}_{j-1}^{+}\} \cap \{\tilde\theta_{jj}>0\}.
$$
The uniqueness of Schur complements of a positive definite matrix also ensures that the encoding--decoding maps are unique for the step $j\to j+1$, and hence, globally (since the $p=1$ case is unique by construction), for both the original matrix and the inverse.

A further consequence of this construction via Schur pivots is the volume of the matrix (the determinant) separates easily. To see this, note that at step $j=p-1$ of Algorithm~\ref{tab:fwd}(a) we define:
\begin{equation}
   \Omegab_{(p-1)\times(p-1)} = \boldsymbol\Theta_{(p-1)\times(p-1)} - \frac{\tthetadotj{p}\tthetadotj{p}^\top}{\tilde\theta_{pp}}.\label{eq:schur}
\end{equation}
Thus, by the Schur determinant formula,
$
\mathrm{det}(\boldsymbol\Theta)= \mathrm{det}(\boldsymbol\Omega_{(p-1)\times(p-1)}) \tilde \theta_{pp}.
$
The recursion in Algorithm~\ref{tab:fwd} then yields:
$$
\mathrm{det(\bTheta)}=\prod_{j=1}^{p}\tilde\theta_{jj},
$$ via an induction argument. 
Thus, we define $v=\sum_{j=1}^{p} \log \tilde\theta_{jj}$ and $d_j = \log\tilde\theta_{jj} - v/p$, so that $\sum_{j=1}^p {d_j}=0,$ and we need only set $\mathbf{d}=(d_2,\ldots,d_{p})^{\top}\in\mathbb{R}^{(p-1)}$, i.e., $d_1$ is not free in our convention.  Finally, we simply define $\mathbf{r}$ to be the vector formed by the upper triangle of $\tilde\bTheta$ above the diagonal, i.e., we have $\mathbf{r}_j=\tilde{\boldsymbol\theta}_{\bullet j}$ for each $j=2,\ldots, p$. The full encoding $\bTheta\mapsto\bx$ consists of  the map $\bTheta\mapsto\tilde\bTheta$ of Algorithm~\ref{tab:fwd} (a), followed by the map $\tilde\bTheta\mapsto\bx$, given as $ \mathbf{x}=(v,\mathbf{d},\mathbf{r})$, where:
\begin{align}
v&=\sum_{j=1}^{p} \log \tilde\theta_{jj},\nonumber\\
d_j &= \log\tilde\theta_{jj} - v/p, \quad j=2,\ldots, p;\; d_1=-\sum_{j=2}^{p} d_j,\nonumber\\
\mathbf{r}_j&=\tilde{\boldsymbol\theta}_{\bullet j}, \quad j=2,\ldots,p.\label{eq:vdr}
\end{align}
Full decoding $\bx\mapsto\bTheta, \bTheta^{-1}$ consists of first mapping $\bx\mapsto\tilde\bTheta$ via backsolving the linear system in~\eqref{eq:vdr} and then applying Algorithm~\ref{tab:fwd}(b) to $\tilde\bTheta$ to decode the original matrix $\bTheta$, or applying Algorithm~\ref{tab:inv}(b) to $\tilde\bTheta$ to decode its inverse $\bTheta^{-1}$. It is clear that $\bx=\mathbf{0}$, where $\mathbf{0}$ is the zero vector of length $p(p+1)/2$, decodes to the identity matrix $\mathbf{I}_p$, representing the origin of the coordinate system. 
%%%%%%%%%%%%%%%%%%%%%%%%%%%%%%%%%%%%%%%%%%%%%%

\subsection{Decoding the Square Roots of the Original Matrix and Its Inverse}
Algorithm~\ref{tab:chol} gives upper triangular $\T$ and lower triangular $\bM$ such that $\T \T^\top=\bTheta$ and $\bM\bM^\top=\bSigma=\bTheta^{-1}.$ Note that decoding $\T$ does not have any sequential dependence over $j=1,\ldots,p$ and the required $\mathcal{O}(p^2)$ calculations can be performed in batch (in contrast to the usual Cholesky algorithm, but one must start from the encoded $\tilde\bTheta$ for this, where encoding has already incurred the $\mathcal{O}(p^3)$ complexity).   The algorithm for $\bM$, in contrast, has sequential dependence and costs $\mathcal{O}(p^3)$. These operations provide valid matrix square roots, in the sense that a draw from $\by\sim \mathcal{N}(0,\bTheta)$ can be performed by drawing $\bz\sim \mathcal{N}(0,\mathbf{I}_p)$ and setting $\by=\T\bz.$ Similarly, a draw from $\by\sim \mathcal{N}(0,\bTheta)$ is obtained as drawing $\bz\sim \mathcal{N}(0,\mathbf{I}_p)$ and  solving $\bM^\top \by=\bz$ via back-substitution \citep[][Chapter 2]{rue2005gaussian}. A crucial property is that $\bM_{j\times j} \bM_{j\times j}^{\top} = \bSigma_{j\times j} = \{\bTheta^{-1}\}_{j\times j}$ for all $j=1,\ldots,p$, whereas $\T_{j:p\times j:p}\T_{j:p\times j:p}^\top=\bTheta_{j:p\times j:p}$, the lower-right $(p-j+1)\times (p-j+1)$ block of $\bTheta$. As a note, although the matrix square roots $\T$ and $\bM$ are a direct consequence of our construction, they appear new, and we are not aware of them being reported in the literature.  
\begin{table}[!h]%
\small
\hspace{0.8cm}
\begin{subalgorithm}{.5\textwidth}
\begin{algorithmic}[1]
 \REQUIRE  $\tilde{\boldsymbol{\Theta}}_{p \times p}$, a symmetric matrix with upper\\ triangle $(\tilde{\boldsymbol\theta}_1, \ldots,\tilde{\boldsymbol\theta}_p)$.
\STATE Initialize $\T_{p \times p} \leftarrow \boldsymbol{0}_{p \times p}$.
\STATE Set $\T_{11}= \tilde\theta_{11}^{1/2}.$
\FOR {($j=2,\ldots, p$)}
\STATE Set $\T_{\bullet j}=\tilde\btheta_{\bullet j}/ \tilde\theta_{jj}^{1/2}.$
\STATE Set $\T_{jj}= \tilde\theta_{jj}^{1/2}.$
%\FOR {($j=p-1,\ldots, 1$)}
\ENDFOR
\ENSURE $\T$, upper triangular, s.t. $\T\T^\top=\bTheta$.
\end{algorithmic}
\caption{Decoding:  $\boldsymbol{\tilde\Theta}_{p\times p}\mapsto\T: \T\T^\top=\bTheta$.}\label{algo_fwd}
\end{subalgorithm}%
%\hspace{2mm}
\begin{subalgorithm}{.5\textwidth}
\begin{algorithmic}[1]

\REQUIRE $\tilde{\boldsymbol{\Theta}}_{p \times p}$, a symmetric matrix with upper \\triangle $(\tilde{\boldsymbol\theta}_1, \ldots,\tilde{\boldsymbol\theta}_p)$.
\STATE Initialize $\bM_{p \times p} \leftarrow \mathbf{0}_{p\times p}.$
\STATE Set $\bM_{11}=\tilde\theta_{11}^{-1/2}$.
\FOR {($j=1,\ldots, p-1$)}
\STATE Set $\bM^{\top}_{\bullet (j+1)}= - \tilde\btheta^{\top}_{\bullet (j+1)}\bM_{j\times j}/\tilde\theta_{j+1,j+1}$.
\STATE Set $\bM_{j+1,j+1}=\tilde\theta_{j+1,j+1}^{-1/2}$.
\STATE Update $\bM_{(j+1) \times (j+1)} \leftarrow \begin{bmatrix}
    {\bM}_{j \times j} & \mathbf{0}\\
    \bM^{\top}_{\bullet (j+1)} & {\bM}_{j+1,j+1}
\end{bmatrix}$
\ENDFOR
 \ENSURE $\bM$, lower triangular, s.t. $\bM\bM^\top=\bTheta^{-1}$.
\end{algorithmic}
\caption{Decoding:  $\boldsymbol{\tilde\Theta}_{p\times p}\mapsto\bM: \bM\bM^\top=\bSigma=\bTheta^{-1}$.}\label{algo_inv}
\end{subalgorithm}%
\captionsetup{labelformat=alglabel}
\caption{Decoding matrix square roots.\\}%
\label{tab:chol}%
\vspace{-0.2cm}
\end{table}
The correctness can be verified via direct computations that compare against Algorithms~\ref{tab:fwd} and~\ref{tab:inv}.

\subsection{Leveraging Sparsity Under Decomposability}
 If the sparsity graph of an SPD matrix is decomposable and the variables are encoded in a perfect elimination ordering, then the RT recursion is zero-fill: the Schur-complement update at each step is a rank-one outer-product update supported on the higher-neighbor set of the eliminated vertex, which is a clique under a perfect elimination ordering. Hence no structural nonzeros are created outside the graph. This follows by the proof of Theorem 9.1 of \citet{vandenberghe2015chordal}, applied to rank-one Schur updates of RT, instead of rank-one Cholesky updates. This yields computational savings under chordal sparsity for the encoder and decoder. However, under arbitrary non-decomposable graphs, fill-ins will generally be created under rank-one adjustments.

\subsection{Jacobians of Transformation}\label{sec:jac}
Note that although it is certainly possible to describe the encoding and decoding maps $\mathbf{x}\leftrightarrow{\bTheta}$ directly, we introduce the intermediate object $\boldsymbol{\tilde{\Theta}}$ to understand its properties of our construction better. To begin, we note from \citet{gao2025order} that a nontrivial  consequence of our construction for $j=p-1,\ldots,1$ is that we have $\tilde{\boldsymbol\theta}_j$  is a function of $(\thetaj{j},\ldots, \thetaj{p})$, of the form:
\begin{align}
\tilde{\boldsymbol\theta}_j &= \thetaj{j} - \phi(\thetaj{j+1},\ldots, \thetaj{p}),\label{eq:nonlinear}
\end{align}
for some nonlinear map $\phi:\mathbb{R}^{(p-j+1)\times j}\to \mathbb{R}^{j}$,
where by convention $\tilde{\boldsymbol\theta}_p={\boldsymbol\theta}_p$. As a consequence it also holds that:
\begin{align}
\tilde{\boldsymbol\theta}_j &= \thetaj{j} - \gamma(\tilde{\boldsymbol\theta}_{j+1},\ldots, \tilde{\boldsymbol\theta}_{p}),\label{eq:nonlinear2}
\end{align}
for another nonlinear map $\gamma:\mathbb{R}^{(p-j+1)\times j}\to \mathbb{R}^{j}$. This linear shift implies:
$$
J(\tilde\btheta_j\mapsto\btheta_j \mid \btheta_{j+1}, \ldots, \btheta_p )=1,
$$
for all $j=1,\ldots, p$, where the case $j=p$ is trivial. As a consequence,
$$
J(\tilde\bTheta_{j:p\times j:p}\mapsto\bTheta_{j:p\times j:p})= \prod_{k=j}^{p} J(\tilde\btheta_k\mapsto\btheta_k \mid \btheta_{k+1}, \ldots, \btheta_p ) =1,
$$ 
for all $j=1,\ldots,p$ and globally,
$$
J(\tilde\bTheta\mapsto\bTheta)=1,
$$ 
as well. Thus, there is volume preservation of the densities under both marginalization and conditioning. To compute probabilities involving the first $j$ variables, one simply drops variables $j+1,\ldots,p$. Sequential addition of variables does not introduce any Jacobian factors in the telescoping chart. This observation was leveraged gainfully by \citet{gao2025order} for developing posterior sampling mechanism in certain Bayesian graphical models without explicit Jacobian corrections, but no further implications as a fully general coordinate system were explored.

The above property also means that for the purpose of density evaluation on the space of SPD matrices $\bTheta$, one can directly evaluate them using $\tilde\bTheta$, which has unrestricted off-diagonal entries and positive diagonal entries (representing the Schur pivots $\tilde\theta_{jj}$), but no SPD constraints. One may also directly evaluate any densities in terms of $\mathbf{x}$. Thus, the Jacobian of transformations satisfy $J(\mathbf{x}\mapsto\bTheta)= J(\mathbf{x}\mapsto\tilde\bTheta)J(\tilde\bTheta\mapsto\bTheta)= J(\mathbf{x}\mapsto\tilde\bTheta)=J((v,\mathbf{d})\mapsto \mathrm{diag}(\tilde\theta_{11}, \ldots, \tilde\theta_{pp}))$, since $\mathbf{r}_j$ is a direct copy of $\tilde\btheta_{\bullet j}$.  Let $s_j=\log\tilde\theta_{jj}=d_j+v/p$. Then, by the chain rule:
\begin{equation}
J(\bx \mapsto \bTheta) = \left| \det \frac{\partial \tilde{\bTheta}}{\partial \bx} \right| = \prod_{j=1}^p \frac{\partial \tilde{\theta}_{jj}}{\partial s_j} = \prod_{j=1}^p \exp(s_j) = \det(\bTheta) = \exp(v),\label{eq:jacob}
\end{equation}
i.e., the Jacobian is determined by single scalar volume coordinate. These properties are considerably simpler compared to Cholesky, matrix logarithm or spectral charts, where there is global entanglement of shape and scale (see Table~\ref{tab:standard-chart-jacobians}). The RT chart isolates all Lebesgue-volume distortion into the scalar log-determinant coordinate. The Jacobian of transformation for $\bSigma=\bTheta^{-1}$ can similarly be computed solely as a function of $v$ as:
\begin{align}
J(\bx\mapsto\bSigma)=J(\bx\mapsto\bTheta)J(\bTheta\mapsto\bSigma)&= \exp(v)\{\det({\bTheta})\}^{-(p+1)}  = \exp(-pv).\label{eq:jacobinv}
\end{align}
The conditional Jacobians also have a simple form:
\begin{align}
J(\tilde\btheta_{ j}\mapsto \bsigma_j \mid \tilde\btheta_1,\ldots\tilde\btheta_{j-1}) &=J(\tilde\btheta_{ j}\mapsto \bsigma_j \mid \bsigma_1,\ldots\bsigma_{j-1})= \{\det(\bSigma_{(j-1)\times(j-1)})\} \tilde{\theta}_{jj}^{-(j+1)}=\left( \prod_{k=1}^{j-1} \tilde{\theta}_{kk} \right)^{-1} \tilde{\theta}_{jj}^{-(j+1)}.\label{eq:JSigma}
\end{align}
Supplementary Section~\ref{sec:jacob} gives a derivation of~\eqref{eq:JSigma}. Combining the expressions in~\eqref{eq:JSigma} also yields:
$$
J(\tilde\bTheta_{j\times j}\mapsto \bSigma_{j\times j})= \left(\prod_{k=1}^{j} \tilde\theta_{kk}\right)^{-(j+1)}.
$$
These expressions are useful for calculating the densities of any functions of $\bSigma$ after decoding the inverse; jointly, marginally and conditionally. The following clarification is now warranted.
\begin{remark}[Rank-one Schur versus rank-one Cholesky updates and Jacobian implications]\label{rem:rtchol}
At this point, it is instructive to briefly highlight the subtle differences between the recursive rank-one Schur updates of \eqref{eq:recursion} and rank-one Cholesky updates in $\bTheta=\bL\D\bL^\top$ parameterization:
\begin{equation}
\boldsymbol\Theta_{j} \leftarrow \underbrace{\begin{bmatrix} \boldsymbol\Theta_{j-1} + \frac{\tilde{\btheta}_{\bullet j}\,\tilde{\btheta}_{\bullet j}^{\top}}{\tilde{\theta}_{jj}} & \tilde{\btheta}_{\bullet j} \\ \tilde{\btheta}_{\bullet j}^\top & \tilde{\theta}_{jj} \end{bmatrix}}_{\text {RT/rank-one Schur update}};\qquad
\boldsymbol\Theta_{j} \leftarrow \underbrace{\begin{bmatrix} \boldsymbol\Theta_{j-1} +  d_{jj}\boldsymbol{\ell}_{\bullet j}\boldsymbol{\ell}_{\bullet j}^{\top} & d_{jj}\boldsymbol{\ell}_{\bullet j}\\ d_{jj}\boldsymbol{\ell}_{\bullet j}^\top & d_{jj} \end{bmatrix}}_{\text{rank-one Cholesky update}}.\label{eq:recursion_chol}
\end{equation}
These two rank-one updates are shown side-by-side in~\eqref{eq:recursion_chol}, and the critical difference is the scaling of the off-diagonal terms by the diagonals for the rank-one Cholesky updates at each $j$ that leads to a global volume--shape entanglement, and the loss of the Jacobian isolation into a single intrinsic scalar coordinate. The RT recursion leaves the off-diagonals unscaled by the diagonals and avoids this entanglement. This is what leads to $J(\tilde\bTheta \mapsto \bTheta)=1$ and $J(\bx\mapsto \bTheta)=\exp(v)$ under RT, but $J((\bL,\D)\mapsto\bTheta)=\prod_{j=1}^{p} d_{jj}^{p-j}$ under Cholesky in Table~\ref{tab:standard-chart-jacobians}. Thus, the Cholesky Jacobian is affected under a permutation of the nodes, due to unequal exponents for $d_{jj}$, but the RT Jacobian is unaffected, since $v$ is invariant under node permutations, despite both Cholesky and RT being triangular maps \citep{marzouk17}.
\end{remark}

\subsubsection{Applications of the Jacobian}

\begin{enumerate}
    \item \textbf{Exact Likelihood Evaluation.} Given a target density $\pi(\bTheta)$ (e.g., Wishart), the density in RT-space is available in closed form via the change-of-variables formula: $\log \pi_{\bX}(\bx) = \log \pi_{\bTheta}(\mathcal{RT}(\bx)) + v$. This enables exact likelihood estimation without numerical Jacobians.
    
    \item \textbf{Integration over the SPD Manifold.} Under the RT transform, integrals over the SPD manifold transform into Euclidean integrals:
    \begin{equation}
    \int_{\mathcal{M}_{p}^{+}} f(\bTheta) d\bTheta = \int_{\mathcal{X}} f(\mathcal{RT}(\bx)) \exp(v) d\bx.
    \end{equation}
    This is particularly useful for computing normalization constants in Bayesian inverse problems.

    \item \textbf{Marginalization and Conditioning.} Marginalization and conditioning contribute unit Jacobian for moving between $\bTheta$ and $\tilde\bTheta$, facilitating computations.

\end{enumerate}

\section{Fisher--Rao Geometry in the Reverse Telescoping Coordinates}\label{sec:fish}
The Fisher--Rao metric on the SPD manifold $\mathcal{M}_{p}^{+}$ is defined by the line element $ds_{FR}^2 = \frac{1}{2} \text{Tr}\left( (\bTheta^{-1} d\bTheta)^2 \right)$ \citep[see][Equation~(6.1), modulo factor $1/2$ scaling convention]{bhatia2009positive}, which is also the standard affine invariant Riemannian metric and the unique Riemannian metric that is invariant under the sufficient statistics \citep{chentsov1982statistical}. We demonstrate that the Reverse Telescoping (RT) chart provides a natural coordinate system for this metric, leading to a recursive, simplified form. First, note from~\eqref{eq:recursion} that the differential $d\bTheta_{j}$ with respect to the $j$-th column coordinates $(\tilde\btheta_{\bullet j}, \tilde{\theta}_{jj})$ is:
\begin{equation*}
d\bTheta_{j} = \begin{bmatrix} -\frac{ \tilde\btheta_{\bullet j}  \tilde\btheta_{\bullet j}^\top}{\tilde{\theta}_{jj}^2} d \tilde{\theta}_{jj} + \frac{1}{\tilde{\theta}_{jj}} (d \tilde\btheta_{\bullet j}  \tilde\btheta_{\bullet j}^\top +  \tilde\btheta_{\bullet j} d \tilde\btheta_{\bullet j}^\top) & d\tilde\btheta_{\bullet j} \\ d\tilde\btheta_{\bullet j}^\top & d \tilde{\theta}_{jj} \end{bmatrix}.
\end{equation*}
The following proposition gives the form of the Fisher--Rao metric.
\begin{proposition}[Fisher--Rao metric in RT coordinates]\label{prop:FR-RT}
The Fisher--Rao metric in RT coordinates admits the additive partitioning:
\begin{equation}
ds^2_{FR} = \underbrace{\frac{1}{2} \sum_{j=1}^{p} (d s_j)^2}_{\text{Volume / Diagonal pivots}} + \underbrace{\sum_{j=2}^{p} \exp(-s_j)\, (D \mathbf{r}_{j})^\top \bSigma_{j-1} \, D\mathbf{r}_{j}}_{\text{Shape (Recursive)}}.\label{eq:fr}
\end{equation}
where,
$ s_j = d_j + v/p= \log \tilde\theta_{jj},\;  \mathbf{r}_{j}= \tilde\btheta_{\bullet j},\; 
D\mathbf{r}_{j}= d \mathbf{r}_{j} - \mathbf{r}_{j} ds_{j}.
$
\end{proposition}
See Supplementary Section~\ref{supp:FR-RT} for a proof of Proposition~\ref{prop:FR-RT}. A critical property of this decomposition is that there is no differential term across $j$. This suggests using the following local whitening where the \emph{past} term $\bSigma_{j-1}$ is frozen, and one sets:
$$
d\z_j =\exp(-s_{j}/2) \bL_{j-1}^{-1} D\mathbf{r}_{j},
$$
so that,
\begin{align*}
    d\z_j^\top d\z_j &=\exp(-s_{j})\, (D \mathbf{r}_{j})^\top (\bL_{j-1}^{-1})^\top\bL_{j-1}^{-1} \, D\mathbf{r}_{j} = \exp(-s_{j})\, (D \mathbf{r}_{j})^\top \bTheta_{j-1}^{-1} \, D\mathbf{r}_{j} = \exp(-s_{j})\, (D \mathbf{r}_{j})^\top \bSigma_{j-1} \, D\mathbf{r}_{j},
\end{align*}
and~\eqref{eq:fr} yields:
$$
ds^2_{FR} = \frac{1}{2} \sum_{j=1}^{p} (d s_j)^2 + \sum_{j=2}^{p} d\z_j^\top d\z_j,
$$
i.e., $(\bz_j, s_j)$ constitute local orthonormal frames. Further, given $\bL_{j-1}$ and $D\r_j$, one can solve for $d\bz_j$  by solving a triangular system with complexity $\mathcal{O}(p^2)$. 

We now contrast this to the Fisher--Rao metric in Cholesky coordinates. We prove the following result, with a proof in Supplementary Section~\ref{supp:frchol}.
\begin{proposition}[Fisher--Rao metric in Cholesky coordinates]\label{prop:frchol}
Let $\bTheta\in \mathcal{M}_{p}^+$ and write its lower-Cholesky factorization as:
\[
\bTheta=\bL \bL^\top,
\qquad
\bL=
\begin{bmatrix}
\bL_{p-1} & 0\\
\bell_p^\top & \alpha_p
\end{bmatrix},
\qquad \alpha_p>0,
\]
and more generally, for each $j=1,\dots,p$,
\[
\bL_j=
\begin{bmatrix}
\bL_{j-1} & 0\\
\bell_j^\top & \alpha_j
\end{bmatrix},
\qquad
\bTheta_j=\bL_j\bL_j^\top,
\qquad
\alpha_j>0,
\]
with the convention that $\bL_1=(\alpha_1)$ and $\bTheta_1=(\alpha_1^2)$. Then the Fisher--Rao line element $ ds_{FR}^2=\sum_j  ds_{FR,j}^2$ admits the recursion:
\begin{align}
 ds_{FR,j}^2
 =ds_{FR,j-1}^2
 +\frac{1}{\alpha_j^2}\bigl\|d\bell_j^\top- \bell_j^\top\bL_{j-1}^{-1} d\bL_{j-1}\bigr\|^2
 +2\left(\frac{d\alpha_j}{\alpha_j}\right)^2.\label{eq:frchol}
\end{align}
\end{proposition}
Comparing the expressions of $D\r_j$ of~\eqref{eq:fr}, which depends only on the current column,  with the differential term $(d\bell_j^\top- \bell_j^\top\bL_{j-1}^{-1} d\bL_{j-1})$ of~\eqref{eq:frchol}, which couples columns $j$ with all previous columns in the differential term through $\bL_{j-1}$, reveals one of the strongest advantages of RT chart. For example, consider the calculation of natural gradients \citep{amari1998natural}. Let $\nabla_\bx \mathcal L$ denote the Euclidean gradient of an objective
$\mathcal L(\bTheta(\bx))$ in RT coordinates. The FR-natural gradient $\widetilde\nabla_\bx \mathcal L$
is defined by:
\[
\widetilde\nabla_\bx \mathcal L \;=\; \G_{\mathrm{FR}}(\bx)^{-1}\nabla_\bx \mathcal L,
\]
where $\G_{\mathrm{FR}}(\bx)$ is the FR metric tensor. The lack of coupling across $j$ under the RT coordinates suggests a natural \emph{splitting} or coordinate-wise computation of gradient over $\bx_j$ with the past columns $1,\ldots, j-1$ held fixed. 
In particular from~\eqref{eq:fr}, for each \(j\ge 2\), the slice \((s_{j},\r_{j})\) has metric block:
\begin{align}
\G_{\mathrm{FR}} (s_j,\r_j)&=
\begin{bmatrix}
\frac12+e^{-s_{j}}\r_{j}^\top\bSigma_{j-1}r_{j}
&
-e^{-s_{j}}\r_{j}^\top\bSigma_{j-1}
\\[1mm]
-e^{-s_{j}}\bSigma_{j-1}\r_{j}
&
e^{-s_{j}}\bSigma_{j-1}
\end{bmatrix},\label{eq:G}
\end{align}
whose inverse is:
\[
\G_{\mathrm{FR}}^{-1}(s_j,\r_j)=
\begin{bmatrix}
2 & 2\r_{j}^\top\\[1mm]
2\r_{j} & e^{s_{j}}\bTheta_{j-1}+2\r_{j}\r_{j}^\top
\end{bmatrix}.
\]
Therefore, if \(\mathcal{L}=\mathcal{L}(s_1,\dots,s_p,\r_2,\dots,\r_{p})\) is a smooth objective, then the exact slice-wise Fisher natural gradient at level \(j\ge 2\) is
\[
\widetilde{\partial_{s_{j}}\mathcal{L}}
=
2\,\partial_{s_{j}}\mathcal{L}+2\,\r_{j}^\top \nabla_{\r_{j}}\mathcal{L},
\]
and,
\[
\widetilde{\nabla_{\r_{j}}\mathcal L}
=
2\,\r_{j}\,\partial_{s_{j}}\mathcal{L}
+
\bigl(e^{s_{j}}\bTheta_{j-1}+2\r_{j}\r_{j}^\top\bigr)\nabla_{\r_{j}}\mathcal{L},
\]
where the per-slice updates of $(s_{j},\r_j)$ given the current $\bTheta_{j-1}$, cost $\mathcal{O}(j^2)$, so that updating over all slices cost $\sum_{j=1}^{p} \mathcal{O}(j^2)=\mathcal{O}(p^3)$.

However, this splitting strategy fails for the Cholesky coordinates, due to the global coupling between columns $j$ with all previous blocks in the differential term, as can be seen from~\eqref{eq:frchol}. To our knowledge, among the standard global SPD parameterizations (e.g. Cholesky/modified Cholesky, spectral or log-based charts), RT is unusual in that its exact Fisher--Rao metric is recursively stage-local: each slice admits a local inverse metric depending only on the current slice variables and the frozen past state, so the exact per-slice natural gradient is $\mathcal{O}(j^2)$, whereas a full sweep remains $\mathcal{O}(p^3)$. By contrast, Cholesky-type charts can be recursively parameterized but not recursively decoupled at the differential level \citep[][Section 3]{lin2019}.

The Fisher--Rao volume form also admits a particularly simple representation, as shown in the next proposition, with a proof in Supplementary Section~\ref{sec:frv}.
\begin{proposition}[Fisher--Rao volume form]\label{prop:frv}
The pulled-back Fisher--Rao volume density is:
\[
d\mathrm{Vol}_{FR}
:=
\sqrt{\det \G_{\mathrm{FR}}(\s,\r)}\,d\s\,d\r
=
2^{-p/2}e^{-(p-1)v/2}\, dv\,d\bd\,d\r =2^{-p/2}(\det\bTheta)^{-(p+1)/2}\,d\bTheta.
\]
\end{proposition}
A remarkable feature of this result is that the volume density depends only on the single scalar $v=\log\det(\bTheta)$ and is insensitive to $(\bd,\r)$. In this way, there is true separation of the determinant of a matrix, encoded by $v$ and the shape of the matrix, encoded by $(\bd,\r)$, under the RT coordinates. Once again a comparison could be drawn with the Cholesky coordinates, where the anisotropic contributions of the diagonal $l_{jj}$ terms prevent separating the volume into a single native chart coordinate. A further useful corollary is that if one defines $\r_{\mathrm{norm}}=\exp(-v/p)\r$, then:
$$
d\mathrm{Vol}_{FR} = 2^{-p/2} dv d\bd d\r_{\mathrm{norm}}.
$$
In Section~\ref{sec:app1}, we will comment further on the implication of this property in designing a split volume--shape flow model, which will be shown to preserve Fisher volume, not just Euclidean volume. The explicit volume density forms may be useful in at least two more ways. First, it provides a closed-form reference measure for Fisher--Rao geometric priors on SPD. Second, it suggests that determinant and shape may be modeled separately even at the level of intrinsic volumes. The final expression in Proposition~\ref{prop:frv} can be seen to coincide with Jeffreys' prior on the covariance matrix $\bTheta$ for multivariate normal model $\mathcal{N}(0,\bTheta)$ \citep{jeffreys1946invariant}, as expected.

\section{The Laplace--Beltrami Operator and Intrinsic Diffusion on SPD}\label{sec:lb}

We present results on some second order properties that are useful for designing intrinsic diffusions on the SPD manifold. In order to do this, first it is helpful to define the following parameterization:
\[
\by=(v,\bd,\bbeta)
=
\bigl(v,d_2,\dots,d_p,\bbeta_2,\dots,\bbeta_p\bigr),
\]
where, $(v,\bd)$ are the same as in $\bx=(v,\bd,\r)$ coordinates, and,
\[
\bbeta_j:=e^{-(d_j+v/p)}\r_j=\frac{\tilde\theta_{\bullet j}}{\tilde\theta_{jj}},\qquad j=2,\dots,p.
\]
The key reason for the reparameterization is that it allows us to avoid mixed derivatives involving $(d_j, \r_j)$ in the subsequent development of this section. For a Riemannian metric \(\G\), the Laplace--Beltrami operator is:
\[
\Delta_{FR}f
=
\frac1{\sqrt{\det \G}}
\partial_i\Bigl(\sqrt{\det \G}\,\G^{ij}\partial_j f\Bigr).
\]
We have the following result, with a proof in Supplementary Section~\ref{sec:laplace}. Along with the desired Laplace--Beltrami operator, it also gives an alternative form of the Fisher--Rao metric.
\begin{proposition}[Fisher--Rao metric and the Laplace--Beltrami operator in $(v,\bd,\bbeta)$ coordinates]
\label{prop:laplace}
The Fisher--Rao line element of ~\eqref{eq:fr} takes the form:
\begin{equation}
\label{eq:metric-vdbeta}
ds_{FR}^2
=
\frac1{2p}dv^2
+
\frac12\sum_{j=1}^p (dd_j)^2
+
\sum_{j=2}^p e^{d_j+v/p}(d\bbeta_j)^\top \bSigma_{j-1}d\bbeta_j,
\end{equation}
giving a block diagonal inverse metric tensor:
\begin{equation}
\label{eq:inverse-metric-vdbeta}
\G_{FR}^{-1}
=
\diag\!\Bigl(
2p,\;
2\P,\;
e^{-(d_2+v/p)}\bTheta_1,\;
\ldots,\;
e^{-(d_p+v/p)}\bTheta_{p-1}
\Bigr),
\end{equation}
where $\P:=\I_{p-1}-\frac1p \mathbf 1 \mathbf 1^\top$. Moreover, the corresponding Laplace--Beltrami operator is:
\begin{equation}
\label{eq:LB-expanded-vdbeta}
\Delta_{FR} f
=
2p\,\partial_{vv}f
+
2\sum_{i,j=2}^p\Bigl(\delta_{ij}-\frac1p\Bigr)\partial_{d_i d_j}f
+
\sum_{j=2}^p (2j-p-1)\,\partial_{d_j}f
+
\sum_{j=2}^p e^{-(d_j+v/p)}\tr\!\bigl(\bTheta_{j-1}\H_{\bbeta_j}f\bigr),
\end{equation}
where \(\H_{\bbeta_j}f\) denotes the Hessian of \(f\) with respect to \(\bbeta_j\) and $\delta_{ij}=\mathbf{1}(i=j)$.
\end{proposition}
\begin{remark}
Under our convention $ds_{FR}^2 = \frac{1}{2} \text{Tr}\left( (\bTheta^{-1} d\bTheta)^2 \right)$, the SPD manifold is known to exhibit sectional curvature $-1\le K\le 0$ \citep[see][Table 5, with $\alpha=1/2,\beta=0$]{thanwerdas2023on}, which is made explicit in coordinates by~\eqref{eq:metric-vdbeta}. If $\bbeta$ is held fixed, the metric is flat in $(v,\bd)$ and attains $K=0$. Conversely, fixing all coordinates except $(d_2,\beta_2)$ gives $ds^2 = dd_2^2 + C e^{2d_2} d\beta_2^2$  for some constant $C>0$, which is the Poincar\'e upper-half-plane metric (with change of variable $y=e^{-d_2})$ and hence attains $K=-1$. 
\end{remark}
The block diagonal structure of $\G^{-1}_{FR}$ in \eqref{eq:inverse-metric-vdbeta} has a direct probabilistic interpretation under a model $\mathcal{N}(0,\bTheta)$. Marginalization over slices $j+1,\ldots, p$ corresponds to dropping the corresponding blocks, contributing no metric distortion. Conditioning on slices $1,\ldots,j-1$ is restricted to the $\bbeta_j$ block $\exp(-(d_j+v/p))\bTheta_{j-1}$, whose geometry is completely determined by the frozen past state $\bTheta_{j-1}$. Thus, the RT coordinates simultaneously simplify both marginal and conditional Fisher–Rao geometry, a property not shared by Cholesky or spectral charts where off-diagonal metric terms couple the conditional and marginal geometries. An immediate corollary of Proposition~\ref{prop:laplace} is the following, with proof in Supplementary Section~\ref{sec:action}.
\begin{corollary}[Action of \(\Delta_{FR}\) on \(v\), \(\bd\), and \(\bbeta\)]
\label{cor:laplace-v-d-beta}
Under the notation of Proposition~\ref{prop:laplace}, the Laplace--Beltrami operator satisfies:
\begin{align*}
\Delta_{FR} v &= 0,\\
\Delta_{FR} \bd_j &= 2j-p-1,\qquad j=2,\dots,p,\\
\Delta_{FR}\bbeta_{j,k} &= 0,\qquad j=2,\dots,p,\ \ k=1,\dots,j-1. 
\end{align*}
Equivalently, \(\Delta_{FR}\bbeta_j=0\) componentwise for every \(j\ge2\).
\end{corollary}
Corollary~\ref{cor:laplace-v-d-beta} shows that the global volume coordinate \(v=\log\det\bTheta\) and the normalized successive regression coordinates \(\bbeta_j\) are harmonic for the Fisher geometry, whereas the determinant-free pivot coordinates \(\bd_j\) have constant Laplacian. Thus, the coordinates \((v,\bd,\bbeta)\) are adapted to the Fisher geometry in a particularly simple way: \(v\) and \(\bbeta\) are harmonic coordinates, while the \(\bd\)-coordinates are affine with constant drift. Nevertheless, we remark here that although the $\by=(v,\bd,\bbeta)$ coordinates may be particularly well-suited for studying the intrinsic geometry under the Fisher--Rao metric, it may not be the best choice for other problems, e.g. density evaluation, as in these coordinates, one loses the remarkably simple Jacobian property $J(\bx\to\bTheta)=\exp(v)$ that holds in the $\bx=(v,\bd,\r)$ coordinates.

Nevertheless, with the intrinsic coordinates thus identified, the natural next step is to specify the intrinsic diffusions in these coordinates. We have the following result, with a proof in Supplementary Section~\ref{sec:diff}.

\begin{lemma}[Intrinsic diffusion on SPD in $(v,\bd,\bbeta)$ coordinates]\label{prop:diff}
Let \(\B=(B_v,B_\bd,B_{\bbeta_2},\ldots,\B_{\bbeta_p})\) be any smooth drift field in \((v,\bd,\bbeta)\) coordinates. Let $\kappa_j=2j-p-1$ with $\boldsymbol{\kappa}=\{\kappa_j\}_{j=2}^{p} \in\mathbb R^{p-1}$, \; $\Q_{j-1}(x)\,\Q_{j-1}(x)^\top=\bTheta_{j-1}(x)$ and $\P=\I_{p-1}-\frac1p \mathbf 1 \mathbf 1^\top$, as before. Then the It\^o diffusion,
\begin{align}
dv_t &= B_v(X_t)\,dt+\sqrt{2p}\,dW_t^{(v)}, \label{eq:gen-v}\\
d\mathbf d_t &= \Bigl(B_{\bd}(X_t)+\frac12\,\boldsymbol{\kappa}\Bigr)\,dt + \sqrt{2}\,\P^{1/2}\,d\W_t^{(d)}, \label{eq:gen-d}\\
d\bbeta_{j,t} &= B_{\bbeta_j}(X_t)\,dt
+ e^{-(d_{j,t}+v_t/p)/2}\,\Q_{j-1}(X_t)\,d\W_t^{(j)},
\qquad j=2,\dots,p, \label{eq:gen-beta}
\end{align}
where \(W^{(v)}\), \(\W^{(d)}\), and \(\W^{(j)}\) are independent standard Brownian motions of dimensions \(1\), \(p-1\), and \(j-1\), respectively, has generator:
\begin{equation}
\label{eq:gen-generator}
Lf = \B\cdot \nabla f + \frac12\Delta_{FR}f.
\end{equation}
In particular, if \(\pi\) is a smooth positive density with respect to Fisher volume \(d\mathrm{Vol}_{FR}\), then the intrinsic Langevin diffusion with invariant measure \(\pi\,d\mathrm{Vol}_{FR}\) is obtained by taking
$
\B=\frac12\,G_{FR}^{-1}\nabla \log\pi.
$
Equivalently, in \((v,\bd,\bbeta)\) coordinates one may write,
\begin{align}
B_v &= p\,\partial_v \log\pi, \label{eq:langevin-Bv}\\
B_d &= \P\,\nabla_{\mathbf d}\log\pi, \label{eq:langevin-Bd}\\
B_{\bbeta_j} &= \frac12\,e^{-(d_j+v/p)}\bTheta_{j-1}\nabla_{\bbeta_j}\log\pi,
\qquad j=2,\dots,p. \label{eq:langevin-Bbeta}
\end{align}
\end{lemma}

\begin{remark}
Lemma~\ref{prop:diff} shows that the intrinsic Brownian motion separates into:
\begin{enumerate}
\item Ordinary Brownian motion in the log-determinant coordinate \(v\),
\item Brownian motion with constant drift on the unit-determinant  hyperplane in the \(\bd\)-coordinates,
\item Drift-free anisotropic Brownian motions in the \(\bbeta_j\)-coordinates, with covariance determined recursively by the frozen-past block \(\bTheta_{j-1}\).
\end{enumerate}
\end{remark}
Similar observations are true for the Langevin diffusion as well. This makes it simple to design intrinsic diffusion, without any explicit accounting for the path Jacobian via numerical calculations. We proceed to discuss two concrete applications of the geometric properties discussed so far in the next two sections.

\section{Application 1: Generative Modeling of SPD-valued Data by Linear Euclidean Split Volume--Shape Flow}\label{sec:app1}
Given two SPD matrices $\bTheta_A$ and $\bTheta_B$, let  $\bx_A=(v_A, \bd_A,\r_A)=\mathcal{T}(\bTheta_A)$ and $\bx_B=(v_B,\bd_B,\r_B)=\mathcal{T}(\bTheta_B)$ be the mapped coordinates via the encoder.  Consider the straight line path joining them:
\begin{align}
\bx(t) &= (1-t) \bx_A + t \bx_B, \qquad t \in [0,1].\label{eq:lindet}
\end{align}
Since the domain of $\bx$ is unconstrained, this defines a valid SPD path, so that $\bTheta(t)=\mathcal{RT}(\bx(t))\in \mathcal{M}^p_+$ uniformly in $t$ with $\det(\bTheta(t))=(\det\bTheta_A)^{1-t}(\det\bTheta_B)^{t}$, since $\det(\bTheta)=\exp(v)$. Thus, along the path we have:
$$
\frac{d}{dt} \log\det(\bTheta_t)=\log\det(\bTheta_B) - \log\det(\bTheta_A),
$$
i.e., log determinant changes linearly along the path. In particular, If $\det(\bTheta_A)=\det(\bTheta_B)=c$, then the entire straight line path joining these two has constant determinant, and suggests designing a flow or diffusion model that maintains a \emph{constant determinant} over the entire flow path.

We leverage this observation to design a transport mechanism for SPD matrices $\bTheta \in \mathcal{M}_{p}^{+}$ that separates global volume from the internal geometry. This is achieved by first normalizing the matrix to the unit-determinant manifold before mapping it to the unconstrained RT-coordinate system.
%\subsection{ Normalization,  Encoding and Solenoidal Flow Design}
Given an SPD matrix $\bTheta(\bx)=\mathcal{RT}(\bx)$ for $\bx=(v,\bd,\r)$ with volume $v = \log\det(\bTheta)$, we define the normalized matrix:
\begin{equation}
\bTheta_{\mathrm{norm}} = \exp(-v/p){\bTheta},\label{eq:vnrom}
\end{equation}
so that $\det(\bTheta_{\mathrm{norm}}) = 1.$ The normalized matrix $\bTheta_{\mathrm{norm}}$ is then encoded via Algorithm~\ref{tab:fwd}(a) into $(0, \mathbf{d}, \boldsymbol{r}_{\mathrm{norm}})$, where $\boldsymbol{r}_{\mathrm{norm}}\in \mathbb{R}^{p(p-1)/2}$ is related to $\mathbf{r}$ as:
\begin{equation}
\boldsymbol{r}_{\mathrm{norm}} = \exp(-v/p) \r.\label{eq:rnorm}
\end{equation}
This allows us to design separate flows for $v$ and $\bxi=(\mathbf{d}, \boldsymbol{r}_{\mathrm{norm}})$, so that $\bx=(0,\bxi)$ always decodes to a unit-determinant matrix, and $v$ and $\bxi\mid v$ can be transported triangularly. The decoded unit-determinant matrix is rescaled at the end of the transport via the transported volume to match the target determinant. The Jacobian of transformation is easily computed as:
$$
J((v,\d,\boldsymbol{r}_{\mathrm{norm}})\mapsto \bTheta)=J((v,\d,\boldsymbol{r}_{\mathrm{norm}})\mapsto \bx)J(\bx\mapsto \bTheta)=\exp\left(\frac{v}{p} \frac{p(p-1)}{2}\right)\exp(v)=\exp\left(\frac{p+1}{2}v\right),
$$
and similarly,
$$
J((v,\d,\boldsymbol{r}_{\mathrm{norm}})\mapsto \bSigma)=\exp\left(-\frac{p+1}{2}v\right).
$$
Using~\eqref{eq:rnorm} in Proposition~\ref{prop:frv} further yields:
\[
d\mathrm{Vol}_{FR}=
2^{-p/2}\, dv\,d\bd\,d\boldsymbol{r}_{\mathrm{norm}} = 2^{-p/2}\, dv\,d\bxi,
\]
suggesting the Fisher--Rao volume density is flat in the $(v,\bxi)$ coordinates, motivating our flow design. 
Given a base distribution $p_0(\mathbf{\bx})$ and target $p_1(\mathbf{\bx})$, we minimize the flow matching objective:
\begin{equation*}
\mathcal{L}_{FM} = \mathbb{E}_{t, q_t(\mathbf{\bx})} \| \mathbf{u}_t(\mathbf{\bx}) - \dot{\bomega}_t(\mathbf{\bx}_0) \|^2,\label{eq:fm}
\end{equation*}
where $\bomega_t(\mathbf{\bx}_0)$ is a probability path. Because the domain of $\mathbf{\bx}$ is the entire Euclidean space $\mathbb{R}^{p(p+1)/2}$, a straight-line path $\bomega_t(\mathbf{\bx}_0) = (1-t)\mathbf{\bx}_0 + t\mathbf{\bx}_1$ always maps to valid SPD matrices under decoding and is the optimal path that minimizes transport cost (with respect to the flat $\mathcal{X}$ space, not geodesically). The target velocity for this path is simply $(\bx_1-\bx_0)$. We train the unit-shape and volume models separately via conditional flow matching \citep{lipman2022flow,albergo2025stochastic}. Given an SPD sample $\bTheta_1$ from the target, we volume-normalize via~\eqref{eq:vnrom}, we compute the pair $(v_1,\bxi_{1})$ and sample $v_0\sim\mathcal{N}(0,1)$.
The training loss for $v$ is defined as:
\begin{align}
\mathcal{L}(v) = \mathbb{E}_{t, v_{0}, v_{1}} \left\| {u}(v_t, t) - (v_1 - v_0) \right\|^2,\label{eq:flowv}
\end{align}
where $v_t=(1-t)v_0+ t v_1$, and sampling proceeds as $dv_t/dt=\mathbf{u}_v(v_t,t)$.
We then sample noise $\bxi_{0} \sim \mathcal{N}(0, \mathbf{I})$ and a random time $t \sim \text{Unif}(0, 1)$. The training loss for $\bxi \mid v=v_1$ is defined as:
\begin{align}
\mathcal{L}({\bxi}\mid v) &= \mathbb{E}_{t, \bxi_{0}, \bxi_{1}, v_1} \left\| \mathbf{u}(\bxi_t,t; v_1) - (\bxi_1 - \bxi_0) \right\|^2,\label{eq:flowxi}
\end{align}
where $\bxi_t = (1-t)\bxi_0 + t\bxi_1$. The probability flow ODE in the coordinate space, conditional on a $v$ sample is: ${d\bxi_t}/{dt} = \mathbf{u}_{\bxi}(\bxi_t,v,t)$. 
The SPD sample reconstructions are as follows.
\begin{enumerate}
    \item \textbf{Reconstruction of $\bTheta$}. The reconstruction follows $\bTheta = \exp(v/p) \mathcal{RT}(\tilde\bTheta)$, by mapping $\bx_t=(0,\bxi)$ into $\tilde\bTheta$ via ~\eqref{eq:vdr}, followed by Algorithm~\ref{tab:fwd}(b) to recover $\mathcal{RT}(\tilde\bTheta)$. %Since $J(\tilde{\bTheta} \to \bTheta) = 1$, the density $p(\bTheta)$ is exactly $p(\tilde{\bTheta})$.
    \item \textbf{Reconstruction of $\bSigma$}. The inverse is recovered via $\bSigma = \exp(-v/p) \mathcal{RT}_{inv}(\tilde\bTheta)$, by mapping $\bx=(0,\bxi)$ into $\tilde\bTheta$ via ~\eqref{eq:vdr}, followed by Algorithm~\ref{tab:inv}(b) to recover $\mathcal{RT}_{inv}(\tilde\bTheta)$.
\end{enumerate}
%If $v$ is transported via a separate 1D diffusion, its stationary density $\pi(v)$ can be modeled independently. 
If $\pi(\bxi \mid v)= \pi(\bd,\boldsymbol{r}_{\mathrm{norm}}\mid v)$ is the terminal density of $\bxi\mid v$ and $\pi(v)$ is the terminal marginal density of $v$ learned by the flow, then the joint terminal density for the SPD matrix and its inverse are:
\begin{equation*}
\pi(\bTheta) = \pi (v)\pi(\bxi\mid v)\exp(-(p+1)v/2), \quad \pi(\bSigma) = \pi(v)\pi(\bxi\mid v)\exp((p+1)v/2).
\end{equation*}
This confirms that the RT-chart allows for joint, marginal, and conditional density evaluations of both $\bTheta$ and $\bSigma$ without complicated Jacobian bookkeeping during the transport process, where the transport of $\bxi$ stays on the unit determinant ($v=0$) path throughout. This can be contrasted with flow matching under log-Cholesky or log-Euclidean pullback \citep{collas2025riemannian}, where straight line paths are possible, but volume--shape split and Jacobian bookkeeping for density evaluation could be nontrivial.

\subsection{Solenoidal Parameterization of the Split Volume--Shape Velocity Field}\label{sec:modham}

Although the Euclidean flow matching of the previous section is simple, and preserves unit-determinant throughout the flow path (if the matrix is normalized before encoding), it is not automatically volume preserving. However, the RT coordinates also make it far simpler to implement a volume-preserving flow compared to most other charts, and we now discuss this. As before, let ${\bxi} = (\mathbf{d}, \mathbf{r}_{\mathrm{norm}}) \in \mathbb{R}^K$ denote the unit-determinant coordinate vector of dimension $K = p(p+1)/2-1$. To strictly enforce the incompressibility of the shape component for the flow matching objectives in ~\eqref{eq:flowv}--\eqref{eq:flowxi}, we parameterize the shape and volume velocity fields $\mathbf{u}(\bxi,t; v)$  and ${u}(v,t)$ separately, where the shape network is indexed by volume.

Specifically, we utilize a Hamiltonian flow model to  enforce the incompressible condition:
\begin{equation*}
\nabla \cdot \mathbf{u}(\bxi,t; v) = 0.
\end{equation*}
Instead of directly predicting $\mathbf{u}(\bxi,t; v)$, we utilize a neural network parameterized by $\psi$ to learn a scalar potential (Hamiltonian) $H_\psi: \mathbb{R}^K \times [0, 1] \times \mathbb{R} \to \mathbb{R}$. The velocity field is then defined as the skew-symmetric gradient of the potential:
\begin{equation}
\mathbf{u}(\bxi,t; v) = \boldsymbol{\Gamma} \nabla_{\bxi} H_\psi(\bxi, t; v),\label{eq:J}
\end{equation}
where $\boldsymbol{\Gamma}=\{\gamma_{ij}\} \in \mathbb{R}^{K \times K}$ is a constant skew-symmetric matrix, satisfying $\boldsymbol{\Gamma} ^\top = -\boldsymbol{\Gamma} $. We choose to model $\boldsymbol{\Gamma}$ via the block-diagonal matrix composed of $2 \times 2$ symplectic blocks:
\begin{equation}
\boldsymbol{\Gamma}  = \text{diag}\left( \begin{bmatrix} 0 & 1 \\ -1 & 0 \end{bmatrix}, \dots, \begin{bmatrix} 0 & 1 \\ -1 & 0 \end{bmatrix} \right).\label{eq:gammatrix}
\end{equation}
If $K$ is odd, we append a one-dimensional zero block. By the properties of skew-symmetric operators, the divergence vanishes identically for any twice-differentiable potential:
\begin{equation*}
\nabla \cdot \mathbf{u}(\bxi,t; v)  = \nabla \cdot (\boldsymbol{\Gamma}  \nabla_{\bxi} H_\psi(\bxi, t; v)) = \sum_{i,j} \gamma_{ij} \frac{\partial^2 H_\psi(\bxi, t; v)}{\partial \bxi_i \partial \bxi_j} = 0.
\end{equation*}
since $\boldsymbol{\Gamma}$ is skew-symmetric and the Hessian matrix is symmetric. This parameterization ensures that the probability path $\bomega_t(0, \bxi)$ remains on the unit-determinant manifold throughout the transport, and the entropy in $\bxi$ is strictly invariant along trajectories. The flow matching objective in ~\eqref{eq:flowxi} is then minimized with respect to the parameters $\psi$ of the potential function $H_\psi$.

For the scalar velocity field ${u}(v,t)$ we use a simple multi-layer perceptron with time features.  All determinant transport is isolated in $v$.

\subsection{Isentropic Properties of the Hamiltonian Flow Model}
We investigate the properties of the Hamiltonian flow model, as specified in Section~\ref{sec:modham}, and show that it is entropy preserving in a precise sense, and with respect to both the Euclidean volume as well as Fisher--Rao volume. Our main result is the following, with a proof in Supplementary Section~\ref{sec:ham}.

\begin{proposition}[Ambient and Fisher--Rao entropy of the flow model]\label{prop:ham}
Let $q_t(\bx_\mathrm{norm})$ be a time-dependent density in the normalized RT coordinates $\bx_{\mathrm{norm}}=(v,\bxi)$, with $\bxi=(\bd,\r_{\mathrm{norm}})$, evolving under the continuity equation
$
 \partial_t q_t + \nabla_{\bx_{\mathrm{norm}}}\cdot(q_t \u)=0,
$
with velocity field of the form: $ \u(\bx_{\mathrm{norm}},t)= (u(v,t),\, \u(\bxi,t; v))$
where $u(v,t)\in \mathbb{R}$ and $\u(\bxi,t; v)\in\mathbb{R}^K$. Assume further that the shape field is Hamiltonian in the shape coordinates, so that:
\[
 \nabla_{\bxi}\cdot \u(\bxi,t; v)\equiv 0.
\]
Then the following hold.

\begin{enumerate}
\item[(i)] If
\[
H_{\bTheta}(q_t)
:= -\int q_t(\bx_{\mathrm{norm}})\log\!\left(\frac{q_t(\bx_{\mathrm{norm}})}{\exp((p+1)v/2)}\right)d\bx_{\mathrm{norm}},
\]
denotes the entropy relative to the ambient SPD volume $d\bTheta$, then,
\[
\frac{d}{dt}H_{\bTheta}(q_t)
=\mathbb E_{q_t}\!\left[\partial_v u(v,t) + \frac{p+1}{2}u(v,t)\right].
\]

\item[(ii)] If
\[
H_{FR}(q_t)
:= -\int q_t(\bx_{\mathrm{norm}})\log\!\left(\frac{q_t(\bx_{\mathrm{norm}})}{2^{-p/2}}\right)d\bx_{\mathrm{norm}},
\]
denotes the entropy relative to the Fisher--Rao volume $d\mathrm{Vol}_{FR}$, then,
\[
\frac{d}{dt}H_{FR}(q_t)
=\mathbb E_{q_t}\!\left[\partial_v u (v,t)\right].
\]
\end{enumerate}
\end{proposition}
A consequence is that under both notions of entropy, the Hamiltonian shape component contributes no entropy production, which is carried only by the determinant drift $u(v,t)$. As a result, if $u(v,t)\equiv 0$, then Proposition~\ref{prop:ham} yields:
\[
\frac{d}{dt}H_{\bTheta}(q_t)=\frac{d}{dt}H_{FR}(q_t)=0.
\]
 Hence, a pure Hamiltonian shape flow is simultaneously entropy-preserving (or isentropic) with respect to both the ambient volume and the Fisher--Rao volume. This has important implications for our design, as a Hamiltonian shape flow for $\bxi\mid v$ on a unit determinant manifold ensures the neural network’s capacity is spent entirely on learning the solenoidal reconfiguration of the shape coordinates, without altering the manifold-valued entropy.

\section{Application 2: Fisher--Rao Distance Bounds Computable in  $\mathcal{O}(p^2)$ on the SPD Manifold}\label{sec:app2}
As another computational application, we show that, under a uniform spectral bound on the SPD matrix, the Fisher--Rao metric in the \((v,\d,\bbeta)\) coordinates is sandwiched between two fixed quadratic forms. This yields an explicit ellipsoidal surrogate for Fisher balls computable in $\mathcal{O}(p^2)$, whereas exact FR distance calculations remain $\mathcal{O}(p^3)$. Recall from \eqref{eq:metric-vdbeta} that in the \((v,\bd,\bbeta)\) coordinates the Fisher--Rao line element is:
\begin{equation*}
ds_{FR}^2
=
\frac{1}{2p}\,dv^2
+
\frac12\sum_{j=1}^p (dd_j)^2
+
\sum_{j=2}^p e^{d_j+v/p}\,(\d\bbeta_j)^\top \bSigma_{j-1}\,\d\bbeta_j.
\end{equation*} 
Then we have the following result, with a proof in Supplementary Section~\ref{app:ellipsoid-bounds}.

\begin{proposition}[Two-sided ellipsoidal bounds under spectral control]
\label{prop:ellipsoid-bounds}
Let \(\bOmega\) be a region in the \((v,\d,\bbeta)\) coordinates such that for every \(\by\in\bOmega\),
\begin{equation}
\label{eq:spectral-control}
m I_p \preceq \Theta(\by) \preceq M I_p
\end{equation}
for some constants \(0<m\le M<\infty\). Then, throughout \(\bOmega\), the Fisher--Rao metric satisfies the two-sided bound:
\begin{equation}
\label{eq:metric-sandwich}
\frac{1}{2p}\,dv^2
+
\frac12\sum_{j=1}^p (dd_j)^2
+
\frac{m}{M}\sum_{j=2}^p \|d\beta_j\|^2
\;\le\;
ds_{FR}^2
\;\le\;
\frac{1}{2p}\,dv^2
+
\frac12\sum_{j=1}^p (dd_j)^2
+
\frac{M}{m}\sum_{j=2}^p \|d\beta_j\|^2.
\end{equation}
\end{proposition}
Proposition~\ref{prop:ellipsoid-bounds} implies that, if
\(\by_0,\by_1\in\bOmega\) and the straight coordinate segment
\[
\gamma(t)=\by_0+t(\by_1-\by_0),\qquad t\in[0,1],
\]
remains in \(\bOmega\), then the restricted Fisher--Rao distance
\[
d_{FR,\bOmega}(\by_0,\by_1)
:=
\inf_{\gamma\subset\bOmega} L_{FR}(\gamma)
\]
satisfies:
\small
\begin{equation}
\label{eq:distance-sandwich}
\frac{(v_1-v_0)^2}{2p}
+
\frac12\,\|\mathbf d^{(1)}-\mathbf d^{(0)}\|^2
+
\frac{m}{M}\sum_{j=2}^p \|\bbeta_j^{(1)}-\bbeta_j^{(0)}\|^2
\le
d^2_{FR,\bOmega}(\by_0,\by_1)
\le 
\frac{(v_1-v_0)^2}{2p}
+
\frac12\,\|\mathbf d^{(1)}-\mathbf d^{(0)}\|^2
+
\frac{M}{m}\sum_{j=2}^p \|\bbeta_j^{(1)}-\bbeta_j^{(0)}\|^2.
\end{equation}
\normalsize
In particular, on \(\bOmega\), the restricted Fisher--Rao distance is
bi-Lipschitz equivalent to an explicit quadratic norm in the
\((v,\bd,\bbeta)\) coordinates. If the spectral bound holds globally on
the full admissible domain, then the same statement holds with
\(d_{FR,\bOmega}\) replaced by the unrestricted distance \(d_{FR}\). Numerical results exploring the tightness of the bounds over varying values of $c=m/M\in (0,1]$ are presented in Section~\ref{sec:app_fish}. The following remarks clarify some implications.

\begin{remark}[Computational complexity]
Evaluating these bounds cost $\mathcal{O}(p^2)$. If $c$ is a global constant on the region of interest, then RT coordinates make Fisher geometry globally equivalent to a weighted Euclidean geometry on $(v,\bd,\bbeta)$ with distortion depending only on $c$. This turns many geometric, optimization and analytic problems on SPD under Fisher geometry to Euclidean questions, up to fixed constants. The penalty is separable in $(v,\bd,\bbeta)$ and the Gaussian quadratic form is $\by^\top\bTheta \by= \sum_{j=1}^{p} e^{s_j}(y_j + \bbeta_j^\top y_{1:j-1})^2$ and $\log\det(\bTheta)=\sum_{j=1}^{p} s_j$. Thus, the per iteration complexity of  Gaussian likelihood evaluation is also $\mathcal{O}(p^2)$.  If $c=m/M\in(0,1]$ is uniform on the search region, the exact Fisher metric and the RT quadratic metric are uniformly equivalent in $(v,\bd,\bbeta)$, with distortion controlled only by $c$.
\end{remark}

\begin{remark}[Ellipsoidal surrogate for Fisher balls and optimization implications]
\label{rem:ellipsoid-surrogate}
Proposition~\ref{prop:ellipsoid-bounds} shows that, on any spectrally controlled region \(\Omega\), Fisher balls are trapped between two explicit ellipsoids in the \((v,\bd,\bbeta)\) coordinates. In particular,
\[
\{\bx:\|\bx-\bx_0\|_{Q_+}\le \rho\}
\subseteq
\{\bx:d_{FR}(\bx,\bx_0)\le \rho\}
\subseteq
\{\bx:\|\bx-\bx_0\|_{Q_-}\le \rho\},
\]
where $Q_{+}$ and $Q_{-}$ are the upper and lower distances. Thus, although the exact Fisher distance is curved and state-dependent, it admits a conservative quadratic surrogate in these coordinates. Suppose one wishes to solve a Fisher-regularized optimization problem of the form:
\[
\min_{\bx\in\Omega}\; F(\bx)+\lambda\,d_{FR}(\bx,\bx_0)^2.
\]
On a spectrally controlled region \(\bOmega\), Proposition~\ref{prop:ellipsoid-bounds} shows that the Fisher penalty can be replaced by explicit quadratic lower and upper surrogates:
\[
\lambda\,\|\bx-\bx_0\|_{Q_-}^2
\;\le\;
\lambda\,d_{FR}(\bx,\bx_0)^2
\;\le\;
\lambda\,\|\bx-\bx_0\|_{Q_+}^2.
\]
This does \emph{not} identify the exact Fisher regularization with a Euclidean quadratic penalty. Rather, it provides a uniform quadratic envelope on \(\bOmega\), which is often sufficient for trust-region design, inner proximal iterations, or conservative local models.
\end{remark}

\setcounter{table}{1}
\section{Numerical Experiments}
\label{sec:num}
We performed basic sanity checks for the encoding and decoding algorithms for the original and the inverse matrix given in Algorithms~\ref{tab:fwd} and~\ref{tab:inv} and their respective Cholesky factors in Algorithm~\ref{tab:chol}. All of these match up to machine precision for all settings tested. Furthermore, the linear relationship for the log-determinant along the straight line path joining $\bx_A=\mathcal{T}(\bTheta_A)$ and $\bx_B=\mathcal{T}(\bTheta_B)$ according to \eqref{eq:lindet}, and that the decoded matrix remains a valid SPD matrix uniformly along the path, are also satisfied up to machine precision. Next, we discuss the results for the two substantive applications discussed in Sections~\ref{sec:app1} and~\ref{sec:app2}, in addition to illustrating an intrinsic Langevin diffusion on the SPD manifold, as specified in Lemma~\ref{prop:diff}. Computer code for synthetic experiments and comparisons with baselines is available on \texttt{GitHub} at: \href{https://github.com/anindyabhadra/RT_SPD}{\nolinkurl{https://github.com/anindyabhadra/RT\_SPD}}.

\subsection{Generative Modeling of SPD-valued Data}\label{sec:gen_app}
\begin{figure}[!t]
\centering
    \includegraphics[width = .99\textwidth,height=7cm]{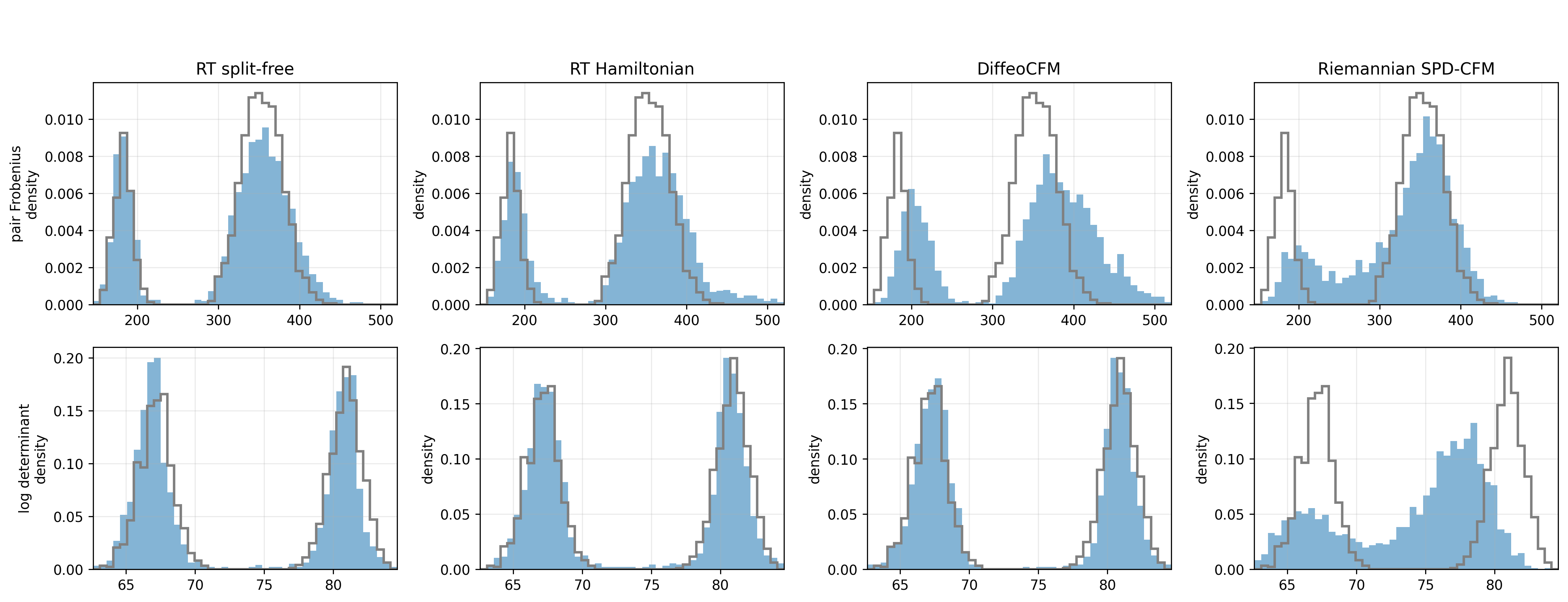}
    \caption{Histogram of pairwise Frobenius distances and log determinant density based on the generated samples for $p=20,\delta=40$ for RT-split flow, RT-Hamiltonian flow, DiffeoCFM and Riemannian CFM. Gray line is based on held out i.i.d. samples from the synthetic target. \label{fig:sim20}}
\end{figure}

\begin{figure}[!t]
\centering
    \includegraphics[width = .99\textwidth,height=7cm]{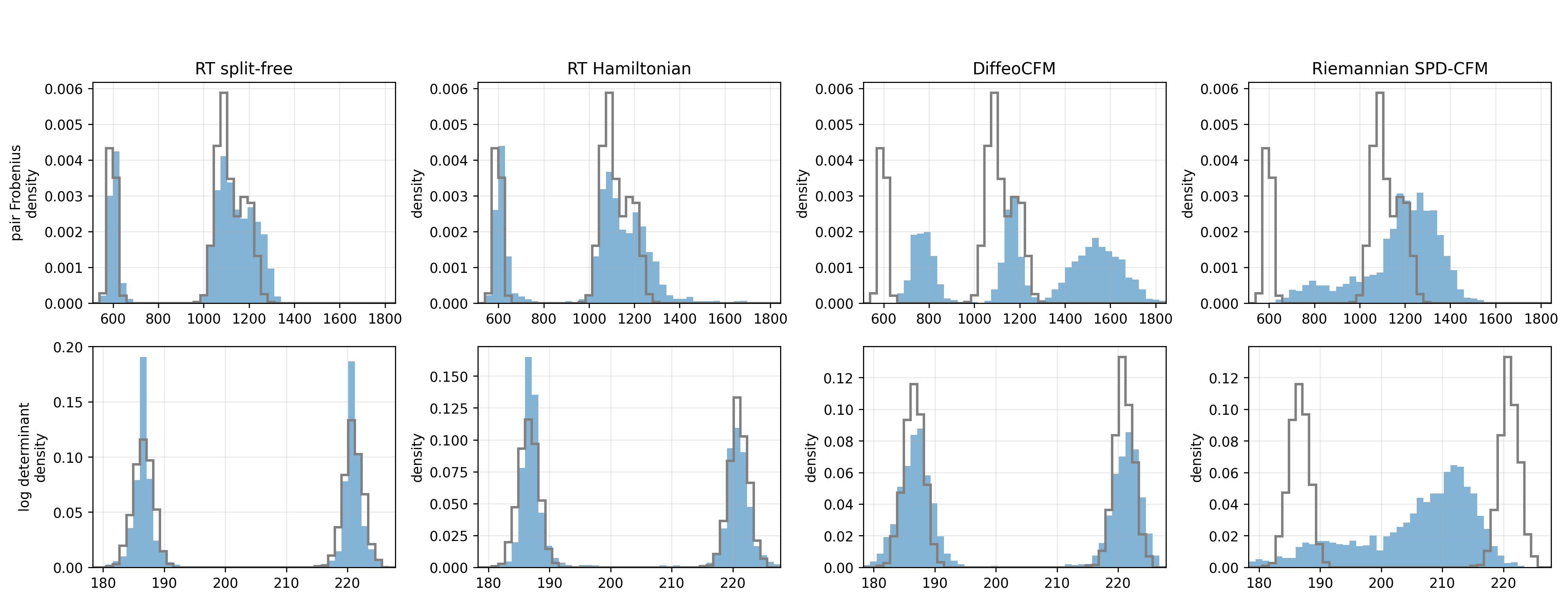}
    \caption{Histogram of pairwise Frobenius distances and log determinant density based on the generated samples for $p=50, \delta=70$ for RT-split flow, RT-Hamiltonian flow, DiffeoCFM and Riemannian CFM. Gray line is based on held out i.i.d. samples from the synthetic target. \label{fig:sim50}}
\end{figure}

\begin{figure}[!h]
\centering
    \includegraphics[width = .99\textwidth,height=7cm]{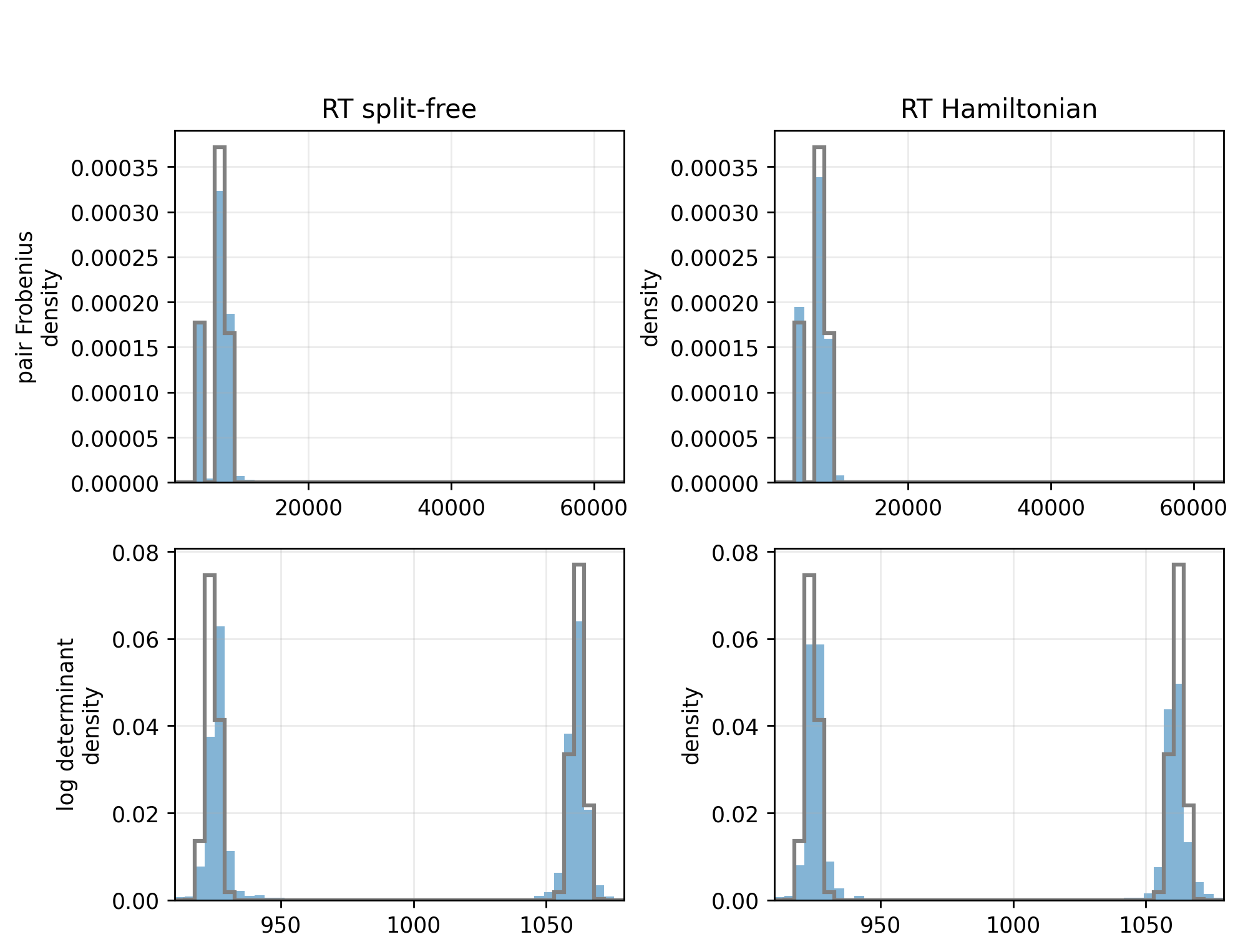}
    \caption{Histogram of pairwise Frobenius distances and log determinant density based on the generated samples for $p=200,\delta=220$ for RT-split flow and RT-Hamiltonian flow. Gray line is based on held out i.i.d. samples from the synthetic target. \label{fig:sim200}}
\end{figure}

We consider a mixture of two Wishart distributions as a synthetic generative target on $\mathcal{M}_p^{+}$:
$$
\pi_{\mathrm{target}}=\frac12 W_p (\delta, \Sigma_1) + \frac12 W_p (\delta, \Sigma_2),
$$
where we set $\Sigma_1$ and $\Sigma_2$ as $p$-dimensional diagonal scale matrices with diagonal values chosen on an equally spaced grid of size $p$ on $[0.8,1.2]$ and $[1.5,2.5]$ respectively, and provide results for $p\in\{20, 50, 100, 200\}$ and $\delta=p+20$. The chosen target is intentionally \emph{hard}, as the distribution is bimodal with the two modes well-separated by a deep valley, where the modes are increasingly spikier as $p$ increases. This is visible in the histograms of one-dimensional log-determinant density we present in Figures~\ref{fig:sim20}--\ref{fig:sim200}, and this is a situation where flow matching is known to struggle due to issues such as flow averaging \citep{guo2025variational,zhang2025towards}.
 
 Given a training sample $\bTheta \in \mathcal{M}_p^{+}$, we follow the volume-normalized encoding and decoding procedures as specified in Section~\ref{sec:app1}, and train the model by minimizing the CFM losses of~\eqref{eq:flowv} and~\eqref{eq:flowxi}. On the split volume--shape domain, we train two types of shape flow: (1) an unrestricted flow model parameterized by multi-layer perceptrons and (2) a divergence free Hamiltonian model. Their specifications are as follows.
 \begin{enumerate}

 \item \textbf{Unconstrained split model.} The model for $v$ is a 3-layer network with SiLU activations functions in the hidden layers, each with 128 hidden nodes. The model for $\bxi\mid v$ is the same, except it uses $\tanh$ in the hidden layers. Optimization for the network parameters is performed using Adam \citep{kingma2014adam}.

\item \textbf{Hamiltonian split model.} The model for $v$ is same as above. The shape coordinate $\bxi \mid v$ is modeled by an exactly divergence-free
Hamiltonian field of ~\eqref{eq:J}, where \(H_{\psi}(\bxi,t; v)\) is a scalar neural Hamiltonian and \(\boldsymbol\Gamma\) is the
canonical symplectic matrix of \eqref{eq:gammatrix}. In the implementation, \(H_{\psi}\) is a
scalar-valued multilayer perceptron with input \((\bxi,t; v)\), two hidden
layers of width \(128\), and $\tanh$ activations. 

\end{enumerate}
Sampling is performed by solving the probability flow ODE \citep{song2020score} using the learned flows and starting from a Gaussian base by successively sampling $v$ and $\bxi\mid v$ as:
\begin{align*}
\frac{dv_t}{dt} &= u(v,t), \qquad v_0 \sim \mathcal{N}(0,1),\\
\frac{d\bxi_t}{dt} &= \u(\bxi,t; v), \qquad \bxi_0 \sim \mathcal{N}(0,\mathbf{I}).
\end{align*}
The standardized volume variable is transported from a standard Gaussian base
using a second-order Heun integrator applied to the learned volume
velocity field. The standardized shape variable is then transported from a
Gaussian base using a second-order Heun integrator applied to
the Hamiltonian shape field, conditional on a generated $v$ sample. The generated
\((v,\bxi)\) samples are then inverse-standardized, recombined into full
RT coordinates, and decoded to SPD matrices.

We use 5,000 training, 2,000 test samples and a holdout set of 2,000 samples drawn directly from the mixture Wishart target. We present results in Figures~\ref{fig:sim20}--\ref{fig:sim200} for $p=20,50,200$, for the two RT-based approaches, and provide comparisons against two benchmarks: DiffeoCFM \citep{collas2025riemannian} under log-Euclidean pullback and Riemannian CFM (RCFM) \citep{chen2024flow}, whenever the latter two were feasible to run (i.e., finished within 1-hr on a CPU).  The figures plot:
\begin{enumerate}
     \item Pairwise Frobenius distance, defined as the $\norm{X_i - Y_i}_F$ for $X_i$ a test SPD sample and $Y_i$ a generated one.

    \item The log determinant density as a one-dimensional summary statistic.
\end{enumerate}


\begin{table*}[!h]
\centering
\small
\begin{tabular}{r l c c c r}
\toprule
$p$ & Method
& MMD
& Sliced $W_2$
& C2ST AUC
& Time (s) \\
\midrule

& Exact holdout
& 0.021 (0.002)
& 0.043 (0.011)
& 0.501 (0.034)
& NA \\

& RT split-free 
& 0.064 (0.001)
& 0.153 (0.091)
& 0.537 (0.05)
& 10.21 (2.12)\\

20
& RT Hamiltonian 
& \textbf{0.028} (0.003)
& \textbf{0.066} (0.005)
& \textbf{0.508} (0.121)
& 11.83 (1.35)\\

& DiffeoCFM
& 0.071 (0.011)
& 0.163 (0.029)
& 0.528 (0.024)
& 35.81 (3.57) \\

& Riemannian SPD-CFM
& 0.077 (0.005) 
& 0.174 (0.022)
& 0.532 (0.051)
& 471.82 (12.32)\\

\addlinespace
\midrule
\addlinespace

& Exact holdout
& 0.011 (0.025)
& 0.041 (0.023)
& 0.501 (0.034)
& NA \\

& RT split-free 
& 0.081 (0.031)
& 0.217 (0.048)
& \textbf{0.514} (0.041)
& 23.42 (3.98)\\

50
& RT Hamiltonian 
& \textbf{0.037} (0.029)
& \textbf{0.082} (0.041)
& 0.532 (0.052)
& 37.32 (3.45) \\

& DiffeoCFM
& 0.121 (0.023) 
& 0.238 (0.057)
& 0.681 (0.198)
& 255.01 (12.12) \\

& Riemannian SPD-CFM
& 0.131 (0.041)
& 0.310 (0.043)
& 0.646 (0.129)
& 2534 (50.97) \\

\addlinespace
\midrule
\addlinespace

& Exact holdout
& 0.013 (0.002)
& 0.051 (0.031)
& 0.507 (0.052)
& NA \\

& RT split-free 
& 0.074 (0.041)
& 0.181 (0.039)
& 0.535 (0.078)
& 85.32 (8.76) \\

100
& RT Hamiltonian 
& \textbf{0.059} (0.021)
& \textbf{0.131} (0.043)
& \textbf{0.518} (0.092)
& 169.30 (17.31)\\

& DiffeoCFM
& --
& --
& --
& -- \\

& Riemannian SPD-CFM
& --
& --
& --
& -- \\
\addlinespace
\midrule
\addlinespace

& Exact holdout
& 0.015 (0.002)
& 0.084 (0.011)
& 0.513 (0.092)
& NA \\

& RT split-free
& 0.090 (0.013)
& 0.210 (0.023)
& 0.531 (0.032)
& 279.60 (25.23)\\

200
& RT Hamiltonian 
& \textbf{0.075} (0.053)
& \textbf{0.147} (0.072)
& \textbf{0.513} (0.079)
& 1625.00 (73.23) \\

& DiffeoCFM
& --
& --
& --
& -- \\

& Riemannian SPD-CFM
& --
& --
& --
& -- \\
\bottomrule
\end{tabular}
\caption{Distributional accuracy and computation time for the Wishart-mixture
experiments. Results report mean
(sd) computed over 20 replicates. MMD and sliced $2$-Wasserstein distance are lower when the
generated and target distributions are closer. For the classifier two-sample
test (C2ST), AUC values closer to $0.5$ indicate that generated and held-out
target samples are difficult to distinguish. The exact-holdout rows compare
two independent samples from the target distribution and provide finite-sample
reference values. Total time includes training and sample generation.
Boldface denotes the best learned method in each distributional-accuracy
column.}
\label{tab:wishart_distributional_metrics}
\end{table*}
At $p=20$, all methods produced reasonable results, but flow averaging under RCFM is noticeable for the log determinant density. DiffeoCFM and RCFM visibly degrade starting around $p=50$. At $p=100$ and $p=200$, RT-based methods are the only feasible ones and performance remains reasonable. Table~\ref{tab:wishart_distributional_metrics} reports the numerical distributional summaries across all settings for the methods and on a holdout set computed from exact i.i.d. samples from the synthetic mixture Wishart target as a controlled benchmark. Maximum mean discrepancy (MMD) under a radial basis function (RBF) kernel \citep{gretton2012kernel} and sliced Wasserstein distances \citep{bonneel2015sliced} are presented as distributional summaries, whereas the areas under curve (AUC) results under a logistic classifier trained by 5-fold cross validation are intended as an adversarial benchmark. Overall, RT Hamiltonian flow performs the best, at the cost of heavier computation needed to compute the scalar Hamiltonian potential for the shape model. Both the RT-based methods outperform the baselines in this example.

\subsection{Fisher--Rao Distance Bound Calculations in $\mathcal{O}(p^2)$}\label{sec:app_fish}
We design a numerical experiment to verify the surrogate Fisher bounds computable in $\mathcal{O}(p^2)$ from \eqref{eq:distance-sandwich} in Section~\ref{sec:app2}. We take $\by_0$ to be the identity matrix and represent a matrix $\bTheta=\bTheta(\by_1)$ in $\by_1=(v_1,\bd_1,\bbeta_1)$ coordinates. Recall:
$$
d^2_{FR}(\mathbf{I},\bTheta) = \frac12 \|\log\bTheta\|_{F}^2.
$$
One can further see from \eqref{eq:distance-sandwich} that the tightness of the bound depends on $c=m/M$. 
\begin{figure}[!t]
\centering
    \includegraphics[width = .9\textwidth]{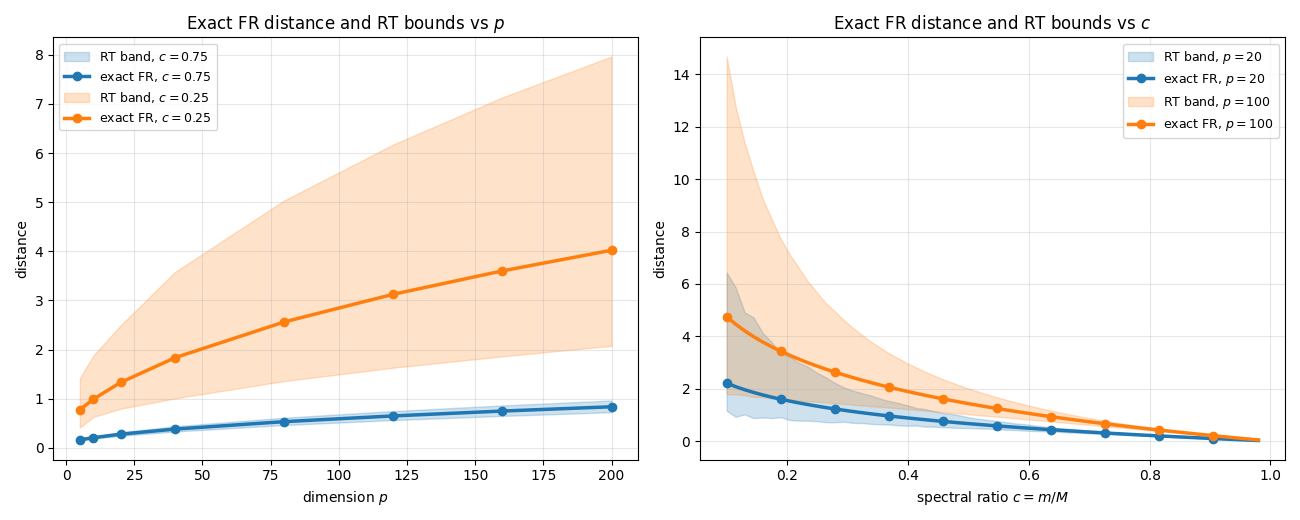}
    \caption{Exact (solid line) and surrogate upper and lower bounds (shaded regions) for FR distance calculations against dimension $p$ (left panel) and spectral ratio $c=m/M$ (right panel).\label{fig:FR}}
\end{figure}
Figure~\ref{fig:FR} displays the exact FR distance and the upper and lower bounds for a unit determinant positive definite matrix against its dimension and spectral ratio. For a well conditioned matrix, the bound remains tight and increasing dimension does not have a noticeable impact, suggesting a good  surrogate to the actual FR distance. That the bounds are computable in $\mathcal{O}(p^2)$ is obvious from \eqref{eq:distance-sandwich} by direct inspection, whereas exact FR distance computation, requires, e.g., an eigenvalue decomposition, which scales as $\mathcal{O}(p^3)$.

\subsection{Intrinsic Langevin Diffusion on the SPD Manifold}\label{sec:app_langevin}
We demonstrate the results of intrinsic Langevin diffusion in $(v,\bd,\bbeta)$ coordinates as specified in Lemma~\ref{prop:diff} under two targets, a single Wishart (which is simple and unimodal) and mixtures of two Wisharts with well separated scale matrices (which is a harder multimodal target):
\begin{align*}
    \pi_1 &=\mathrm{W}_{p}\left(\delta, \frac{1}{\delta}I_p\right),\\
    \pi_2 &= \frac12 \mathrm{W}_{p}(\delta, I_p) + \frac12 \mathrm{W}_{p}\left(\delta, I_p + k\frac{\mathbf{1}\mathbf{1}^\top}{p}\right).
\end{align*}
We set $\delta=p+20$ and $k=4$. We run 10,000 unadjusted intrinsic Langevin iterations with the first 3,000 discarded as burn-in with a step size of 2e-5. Although this algorithm does not in general leave the target distribution invariant at stationarity without a Metropolis correction, performance guarantees are available \citep{durmus2019high} and we choose this as a simple approach for demonstrating intrinsic diffusion. The required gradients are analytically computable for our synthetic targets, but this need not be the case in general, and consequently we use automatic differentiation for gradient evaluation. 
\begin{figure}[!t]
\centering
    \includegraphics[width = .95\textwidth]{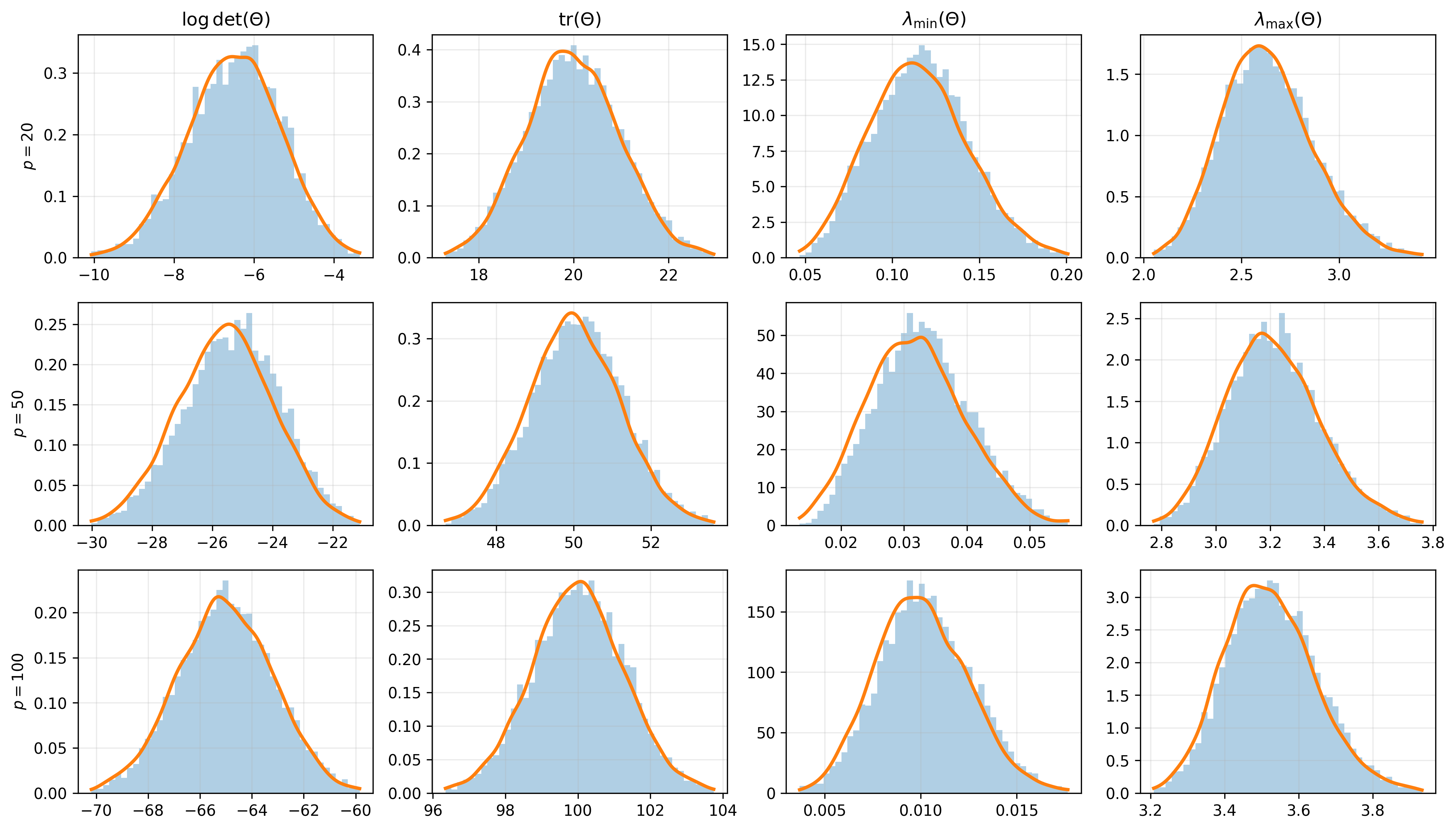}
    \caption{Empirical histograms of (from left) log-determinant, trace, minimum and maximum eigenvalues of the generated SPD samples under an intrinsic Langevin diffusion targeting $\pi_1$ for $p=20,50,100$. The solid orange line in each plot is generated by kernel density estimate under i.i.d. samples from the same Wishart distribution. \label{fig:Lang1}}
\end{figure}

\begin{figure}[!h]
\centering
    \includegraphics[width = .95\textwidth]{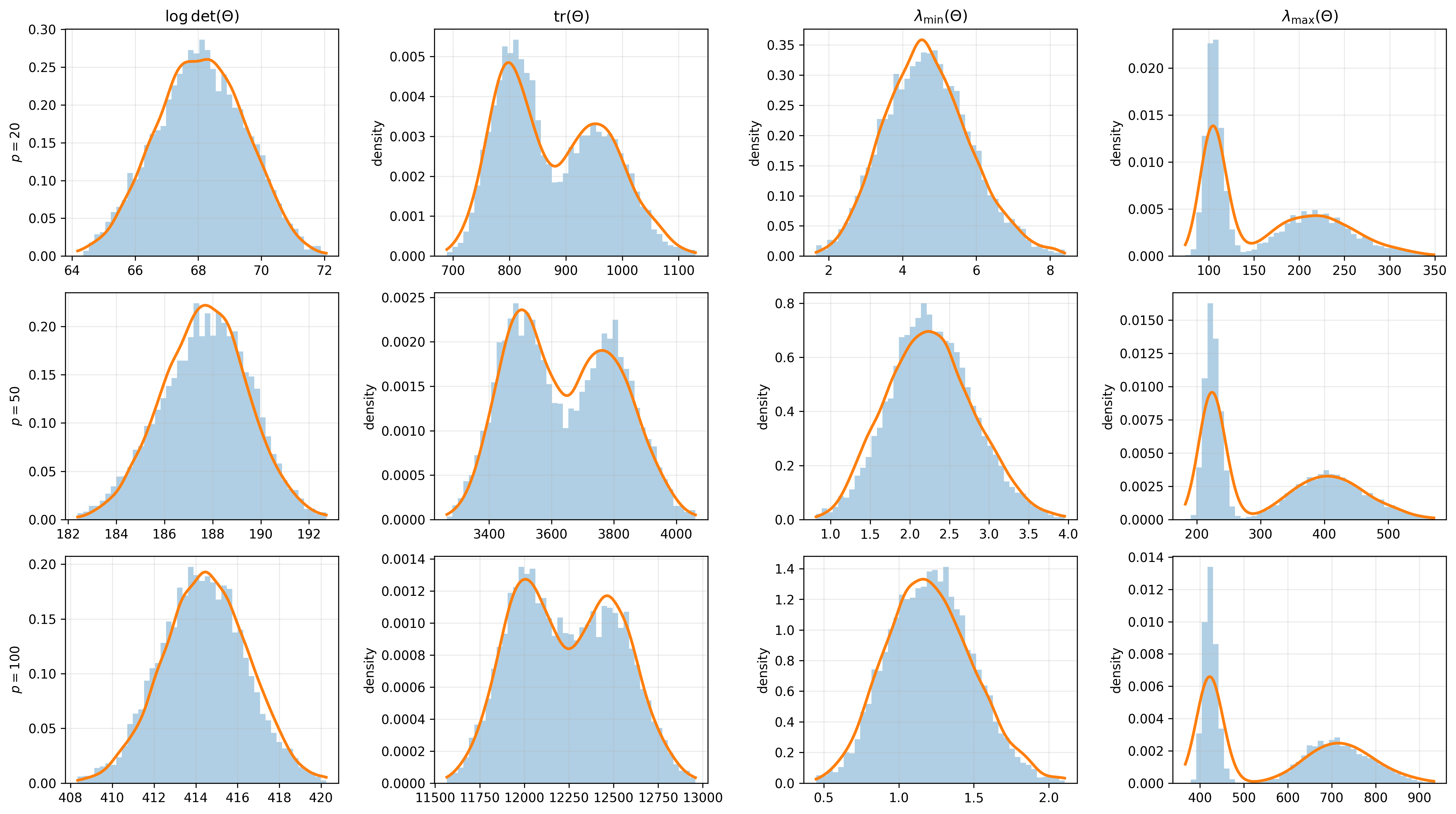}
    \caption{Empirical histograms of (from left) log-determinant, trace, minimum and maximum eigenvalues of the generated SPD samples under an intrinsic Langevin diffusion targeting $\pi_2$ for $p=20,50,100$. The solid orange line in each plot is generated by kernel density estimate under i.i.d. samples from the same mixture Wishart distribution. \label{fig:Lang2}}
\end{figure}
We present results for $p=20,50,100$ in Figures~\ref{fig:Lang1} and~\ref{fig:Lang2}, where we present histograms based on generated samples for  log-determinant, trace, minimum and maximum eigenvalues of the generated SPD samples. As a reference point, kernel density estimate based on 8,000 direct samples from each of these distributions are also presented as the corresponding truth. As expected, results are better when the target is $\pi_1$ as $\pi_2$ is a highly bimodal target, which is noticeable especially for the trace and the maximum eigenvalue distributions. However, performance remains reasonable, and no SPD-specific restriction is needed to implement the samplers in the unconstrained intrinsic coordinates. These results complement the generative results in Section~\ref{sec:gen_app}, and show the proposed approach is useful for SPD-constrained Bayesian inverse problems as well, where the goal is to sample from a given target distribution, typically a Bayesian posterior. 

\section{Generative Modeling of Brain Functional Connectivity Networks}\label{sec:brain}
\begin{figure}[!h]
\centering
    \includegraphics[width = .95\textwidth]{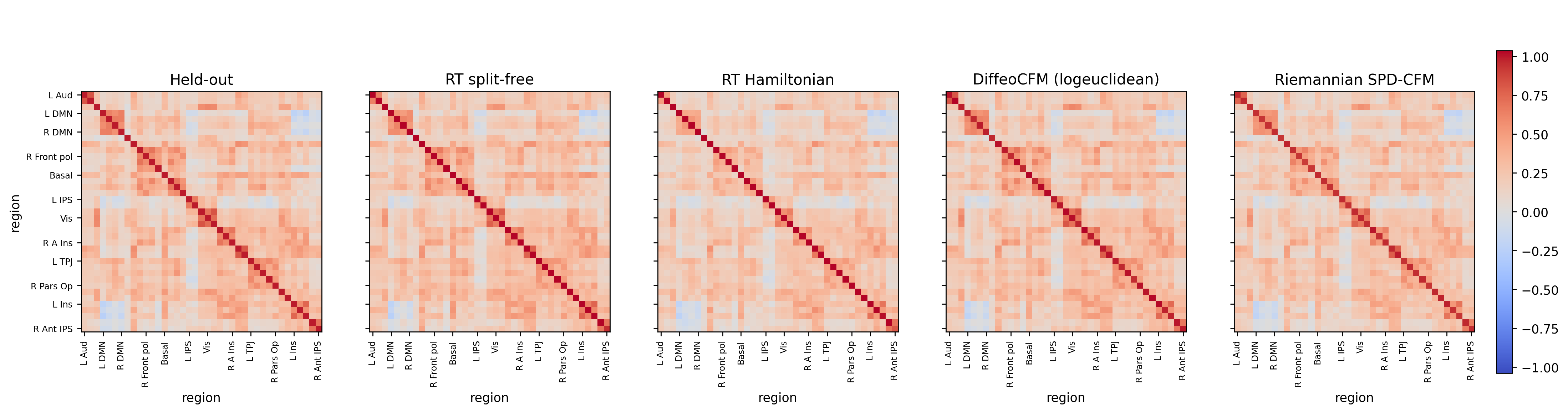}
    \caption{Median ABIDE functional connectivity heatmaps for held-out subjects and samples generated by RT split-free, RT Hamiltonian, DiffeoCFM, and Riemannian SPD-CFM. All panels use the same color scale.\label{fig:abide}}
\end{figure}

%We present results on two brain imaging applications.
%\subsection{The ABIDE Study}
We obtained the preprocessed  Autism Brain Imaging Data Exchange (ABIDE) functional magnetic resonance imaging (fMRI) data \citep{nielsen2013abide} using the Nilearn interface (\href{https://nilearn.github.io/dev/modules/description/ABIDE_pcp.html}{\nolinkurl{https://nilearn.github.io/dev/modules/description/ABIDE\_pcp.html}}), consisting of both individuals with autism spectrum disorder (ASD) and typical controls. Regional time series were extracted using the Multi-subject Dictionary Learning (MSDL) atlas, which gives $p=39$ functional regions, and subject-level $39\times 39$ functional connectivity matrices were formed and treated as SPD-valued observations. After preprocessing and filtering, the data set contained $n=871$ individuals. We use 20 random splits of the data set into training and held-out samples with proportions 80\% and 20\% in each split, and train the RT based methods using the models as described in Section~\ref{sec:gen_app}. We also compare against DiffeoCFM \citep{collas2025riemannian} under a log-Euclidean pullback and Riemannian SPD-CFM \citep{chen2024flow}. Figure~\ref{fig:abide} plots the median connectivity heatmaps for empirical holdout and generated samples. Visually, all methods  capture the dominant patterns in the data that emerge from the holdout samples. Table~\ref{tab:abide} reports the distributional summary and computation times. RT Hamiltonian gives the best C2ST AUC and sliced-Wasserstein summaries among the learned methods; while Riemannian SPD-CFM gives the best MMD, although the differences are small relative to standard deviations across splits for MMD and AUC. 

\begin{table*}[!h]
\centering
\small
\begin{tabular}{l c c c r}
\toprule
 Method
& MMD
& Sliced $W_2$
& C2ST AUC
& Time (s) \\
\midrule

Empirical holdout
& 0.092 (0.023)
& 0.232 (0.031)
& 0.549 (0.039)
& NA \\

 RT split-free 
& 0.091 (0.014)
& 0.224 (0.030)
& 0.540 (0.031)
& 24.16 (2.26) \\

 RT Hamiltonian 
& 0.090 (0.017)
& \textbf{0.208} (0.037)
& \textbf{0.530} (0.020)
& 27.18 (2.11) \\

 DiffeoCFM
& 0.098 (0.015)
& 0.343 (0.190)
& 0.537 (0.027)
& 41.35 (4.73) \\

 Riemannian SPD-CFM
& \textbf{0.089} (0.017)
& 0.219 (0.026)
& 0.559 (0.034)
& 361.84 (11.1)\\
\bottomrule
\end{tabular}
\caption{Distributional accuracy and computation time for the ABIDE data. Results report mean (sd) computed over 20 random training--test splits. MMD and sliced $2$-Wasserstein distance are lower when the
generated and target distributions are closer. For the classifier two-sample
test (C2ST), AUC values closer to $0.5$ indicate that generated and held-out
target samples are difficult to distinguish. The empirical holdout rows compare
two independent samples based on empirical splits of the data and provide reference values. Total time includes training and sample generation.
Boldface denotes the best learned method in each
column.}
\label{tab:abide}
\end{table*}

\section{Conclusions}
We propose a new unconstrained coordinate system for SPD matrices that demonstrates nontrivial demarcations from existing unconstrained charts such as matrix log, Cholesky or spectral charts. The most salient feature is the isolation of the determinant and the Jacobian in a single scalar coordinate, which is a demarcation from other existing charts. This leads to several nontrivial geometric and algorithmic consequences that we thoroughly explore in the rest of the paper, including intrinsic diffusion and split volume--shape generative flow models. 

The present paper explores the RT coordinates and their implications from several angles, but of course, it is not possible to explore or even anticipate all consequences in a single paper. Nevertheless, some areas bear mentioning. For generative modeling, we explore approaches based on CFM due to their speed and flexibility. Nevertheless, score-based diffusion models may be preferable for generative tasks if sample diversity in CFM becomes a concern. Similar to intrinsic sampling from a target distribution described in Section~\ref{sec:app_langevin}, we expect the RT chart to enable intrinsic diffusion-based generative modeling in a manner far easier than most other charts. On the other hand, one may also explore sharper geometric consequences, such as Levi--Civita connections or curvature computation in RT coordinates. We leave these areas open for future investigations.

\section*{Supplementary Material}
The Supplementary Material contains all technical proofs. Computer code with complete documentation for the synthetic experiments is freely available on \texttt{GitHub} at URL: \href{https://github.com/anindyabhadra/RT_SPD}{\nolinkurl{https://github.com/anindyabhadra/RT\_SPD}}.

\bibliographystyle{apalike}
\bibliography{hs-review, ref}

\clearpage\pagebreak\newpage
%%%%%%%%%%%%%%%%%%%%%%%%%%%%%%%%%%%%%%%%%%%%%%
\begin{center}
	{\LARGE{\bf Supplementary Material to\\
	{\it The Reverse Telescoping Coordinate System for Positive Definite
Matrices: Geometry, Computation, and Generative Modeling}}
	}
\end{center}
\setcounter{equation}{0}
\setcounter{page}{1}
\setcounter{table}{0}
\setcounter{section}{0}
\setcounter{subsection}{0}
\setcounter{figure}{0}
\renewcommand{\theequation}{S.\arabic{equation}}
\renewcommand{\thesection}{S.\arabic{section}}
\renewcommand{\thepage}{S.\arabic{page}}
\renewcommand{\thetable}{S.\arabic{table}}
\renewcommand{\thefigure}{S.\arabic{figure}}
%%%%%%%%%%%%%%%%%%%%%%%%%%%%%%%%%%%%%%%%%%%%%%%%%%%%
%%%%%%%%%%%%%%%%%%%%%%%%%%%%%%%%%%%%%%%%%%%%%%%%%%%%%%%%%%
%\section*{Appendix}
%\appendix
\section{Derivation of~\eqref{eq:JSigma}}\label{sec:jacob}
We seek the conditional Jacobian of the $j$-th step, representing the change in the new precision variables $(\mathbf{v}_j, \sigma_{jj})$ with respect to the $j$-th column coordinates $(\tilde\btheta_{\bullet j}, \tilde\theta_{jj})$. Below, we abbreviate $\mathbf{u}_j = \tilde\btheta_{\bullet j}$ and $d_j = \tilde{\theta}_{jj}$.
From~\eqref{eq:recursion}, the $j$-th block is defined by:
\begin{align*}
\mathbf{v}_j &= -\bSigma_{j-1} \left( \frac{\mathbf{u_j}}{d_j} \right), \\
\sigma_{jj} &= \frac{1}{d_j} + \frac{\mathbf{u}_j^\top \bSigma_{j-1} \mathbf{u}_j}{d_j^2}.
\end{align*}
The Jacobian matrix $J_j$ for this conditional mapping is:
\begin{equation*}
J_j = \frac{\partial(\mathbf{v}_j, \sigma_{jj})}{\partial(\mathbf{u}_j, d_j)} = 
\begin{bmatrix} 
-d_j^{-1} \bSigma_{j-1} & d_j^{-2} \bSigma_{j-1} \mathbf{u}_j \\ 
2 d_j^{-2} \mathbf{u}_j^\top \bSigma_{j-1} & -d_j^{-2} - 2 d_j^{-3} \mathbf{u}_j^\top \bSigma_{j-1} \mathbf{u}_j
\end{bmatrix}.
\end{equation*}
Using the Schur determinant formula: $\det \begin{bmatrix} A & B \\ C & D \end{bmatrix} = \det(A)\det(D - CA^{-1}B)$, we compute:
\begin{align*}
    \det(A) &= \det(-d_j^{-1} \bSigma_{j-1}) = (-d_j^{-1})^{j-1} \det(\bSigma_{j-1}),\\
    \det(D-CA^{-1}B) &= (-d_j^{-2} - 2d_j^{-3} \mathbf{u}_j^\top \bSigma_{j-1} \mathbf{u}_j) - (2d_j^{-2} \mathbf{u}_j^\top \bSigma_{j-1}) (-d_j \bSigma_{j-1}^{-1}) (d_j^{-2} \bSigma_{j-1} \mathbf{u}_j)\\
     &= -d_j^{-2} - 2d_j^{-3} \mathbf{u}_j^\top \bSigma_{j-1} \mathbf{u}_j + 2d_j^{-3} \mathbf{u}_j^\top \bSigma_{j-1} \mathbf{u}_j = -d_j^{-2}.
\end{align*}
The conditional Jacobian magnitude is:
\begin{equation*}
|J_j| = \left| (-d_j^{-1})^{j-1} \det(\bSigma_{j-1}) (-d_j^{-2}) \right| = d_j^{-(j+1)} \det(\bSigma_{j-1}).
\end{equation*}
Since $\det(\bSigma_{j-1})$ is the reciprocal of the determinant of the $j-1$ covariance sub-block, it is expressed as the product of the preceding pivots:
\begin{equation*}
\det(\bSigma_{j-1}) = \left( \prod_{k=1}^{j-1} \tilde{\theta}_{kk} \right)^{-1}.
\end{equation*}
Substituting this into $|J_j|$, we obtain the final closed-form expression:
\begin{equation*}
\left| \frac{\partial (\mathbf{v}_j, \sigma_{jj})}{\partial (\tilde\btheta_{\bullet j}, \tilde{\theta}_{jj})} \right| = \left( \prod_{k=1}^{j-1} \tilde{\theta}_{kk} \right)^{-1} \tilde{\theta}_{jj}^{-(j+1)}.
\end{equation*}

\section{Proof of Proposition~\ref{prop:FR-RT}}\label{supp:FR-RT}
The Fisher--Rao (FR) metric on the SPD manifold $\mathcal M_p^+$ is defined by the line element:
\begin{equation*}
  ds^2_{\mathrm{FR}}(\bTheta)
  = \frac12 \, \mathrm{Tr}\!\left\{ (\bTheta^{-1} d\bTheta)^2 \right\}
  = \frac12 \, \mathrm{Tr}\!\left\{ (\bSigma \, d\bTheta)^2 \right\},
\end{equation*}
where $\bSigma = \bTheta^{-1}$. 
We define the \emph{incremental} telescoping contribution of level $j$ to the FR line element by:
\begin{equation*}
  ds^2_{\mathrm{FR},j} (\bTheta)
  \;:=\;\frac12\mathrm{Tr}\!\left\{\big(\bSigma_{j}\,d\bTheta_{j}\big)^2\right\}
      -\frac12\mathrm{Tr}\!\left\{\big(\bSigma_{j-1}\,d\bTheta_{j-1}\big)^2\right\},
\end{equation*}
so that $ds^2_{\mathrm{FR}}(\bTheta)= \sum_{j=1}^{p}ds^2_{\mathrm{FR},j}(\bTheta)$. 
To prove the result, we define:
\begin{equation*}
  \bbeta_{\bullet j} := \frac{\tilde\btheta_{\bullet j}}{\tilde\theta_{jj}}.
\end{equation*}
and recall the block partition formula from ~\eqref{eq:recursion} for $j=2,\dots,p$:
\begin{equation*}%\label{eq:FR-recursion-Theta-Sigma}
\bTheta_{j} =
\begin{bmatrix}
\bTheta_{j-1} + \tilde\theta_{jj}\,\bbeta_{\bullet j}\bbeta_{\bullet j}^\top & \tilde\theta_{jj}\,\bbeta_{\bullet j} \\
\tilde\theta_{jj}\,\bbeta_{\bullet j}^\top & \tilde\theta_{jj}
\end{bmatrix},
\;
\bSigma_{j} =
\begin{bmatrix}
\bSigma_{j-1} & -\bSigma_{j-1}\bbeta_{\bullet j} \\
-\bbeta_{\bullet j}^\top \bSigma_{j-1} & \tilde\theta_{jj}^{-1} + \bbeta_{\bullet j}^\top \bSigma_{j-1} \bbeta_{\bullet j}
\end{bmatrix}.
\end{equation*}
We abbreviate:
\begin{equation*}
  a := \tilde\theta_{jj},
  \qquad
  \bbeta := \bbeta_{\bullet j},
  \qquad
  \S := \bSigma_{j-1} = \bTheta_{j-1}^{-1},
\end{equation*}
so that,
\begin{equation*}
\bTheta_{j} =
\begin{bmatrix}
\bTheta_{j-1} + a\bbeta\bbeta^\top & a\bbeta \\
a\bbeta^\top & a
\end{bmatrix},
\qquad
\bSigma_{j} =
\begin{bmatrix}
\S & -\S\bbeta \\
-\bbeta^\top \S & a^{-1}+\bbeta^\top \S\bbeta
\end{bmatrix}.
\end{equation*}
Throughout the derivation below, the predecessor block $(\bTheta_{j-1},\bSigma_{j-1})$ is held fixed while differentiating the new coordinates $(\bbeta,a)$.
%\paragraph{Step 1: Differential of $\Theta_{j+1}$.}
Differentiating the block representation gives:
\begin{equation*}
  d\bTheta_{j} =
  \begin{bmatrix}
    d(a\bbeta\bbeta^\top) & d(a\bbeta) \\
    d(a\bbeta^\top) & da
  \end{bmatrix}.
\end{equation*}
Using the chain rule,
\begin{equation*}
  d(a\bbeta\bbeta^\top)
  = \bbeta\bbeta^\top \, da + a\,(d\bbeta\,\bbeta^\top + \bbeta\,d\bbeta^\top),
  \qquad
  d(a\bbeta)=\bbeta\,da + a\,d\bbeta.
\end{equation*}
Thus,
\begin{equation*}
  d\bTheta_{j}
  =
  \begin{bmatrix}
    \bbeta\bbeta^\top \, da + a(d\bbeta\,\bbeta^\top + \bbeta\,d\bbeta^\top)
    & \bbeta\,da + a\,d\bbeta \\
    \bbeta^\top da + a\,d\bbeta^\top & da
  \end{bmatrix}.
\end{equation*}
%\paragraph{Step 2: Compute $\Sigma_{j+1}d\Theta_{j+1}$.}
We further simplify notations by writing:
\begin{equation*}
  d\bTheta_{j} = \begin{bmatrix} \A & \b \\ \b^\top & da \end{bmatrix},
\end{equation*}
where,
\begin{equation*}
  \A = \bbeta\bbeta^\top da + a(d\bbeta\,\bbeta^\top + \bbeta\,d\bbeta^\top),
  \qquad
  \b = \bbeta\,da + a\,d\bbeta.
\end{equation*}
Then,
\begin{equation*}
  \bM_{j}:=\bSigma_{j}d\bTheta_{j}
  =
  \begin{bmatrix}
    \S\A-\S\bbeta \b^\top & \S\b-\S\bbeta da \\
    -\bbeta^\top \S\A + (a^{-1}+\bbeta^\top \S\bbeta)b^\top & -\bbeta^\top \S\b + (a^{-1}+\bbeta^\top \S\bbeta)da
  \end{bmatrix}.
\end{equation*}
We simplify each block separately.
\begin{enumerate}
\item For the upper-left block,
\begin{align*}
\S\A - \S\bbeta \b^\top
&= \Bigl[\S\bbeta\bbeta^\top da + a\S\,d\bbeta\,\bbeta^\top + a\S\bbeta\,d\bbeta^\top\Bigr]
   - \Bigl[\S\bbeta(\bbeta^\top da + a\,d\bbeta^\top)\Bigr] \\
&= a\S\,d\bbeta\,\bbeta^\top.
\end{align*}
\item For the upper-right block,
\begin{align*}
\S\b - \S\bbeta da
&= \S(\bbeta da + a\,d\bbeta) - \S\bbeta da \\
&= a\S\,d\bbeta.
\end{align*}
\item For the lower-left block,
\begin{align*}
&-\bbeta^\top \S\A + (a^{-1}+\bbeta^\top \S\bbeta)\b^\top \\
&= -\bbeta^\top\Bigl[\S\bbeta\bbeta^\top da + a\S\,d\bbeta\,\bbeta^\top + a\S\bbeta\,d\bbeta^\top\Bigr]
   + (a^{-1}+\bbeta^\top \S\bbeta)(\bbeta^\top da + a\,d\bbeta^\top) \\
&= - (\bbeta^\top \S\bbeta)\bbeta^\top da - a\,\bbeta^\top \S d\bbeta\,\bbeta^\top - a(\bbeta^\top \S\bbeta)d\bbeta^\top \\
&\qquad + a^{-1}\bbeta^\top da + (\bbeta^\top \S\bbeta)\bbeta^\top da + d\bbeta^\top + a(\bbeta^\top \S\bbeta)d\bbeta^\top \\
&= d\bbeta^\top + \frac{\bbeta^\top}{a}da - a(\bbeta^\top \S d\bbeta)\bbeta^\top.
\end{align*}
\item Finally, for the lower-right block,
\begin{align*}
-\bbeta^\top \S\b + (a^{-1}+\bbeta^\top \S\bbeta)da
&= -\bbeta^\top \S(\bbeta da + a\,d\bbeta) + (a^{-1}+\bbeta^\top \S\bbeta)da \\
&= -(\bbeta^\top \S\bbeta)da - a\,\bbeta^\top \S d\bbeta + a^{-1}da + (\bbeta^\top \S\bbeta)da \\
&= \frac{da}{a} - a\,\bbeta^\top \S d\bbeta.
\end{align*}
\end{enumerate}
Therefore,
\begin{equation*}
  \bM_{j} = \bSigma_{j}d\bTheta_{j}
  =
  \begin{bmatrix}
    a\S\,d\bbeta\,\bbeta^\top & a\S\,d\bbeta \\
    d\bbeta^\top + \dfrac{\bbeta^\top}{a}da - a(\bbeta^\top \S d\bbeta)\bbeta^\top
    & \dfrac{da}{a} - a\,\bbeta^\top \S d\bbeta
  \end{bmatrix}.
\end{equation*}
We proceed to compute the trace of the square. Introduce the abbreviations:
\begin{equation*}
  \bx := \a\S\,d\bbeta,
  \qquad
  t := \bbeta^\top \bx = a\,\bbeta^\top \S d\bbeta,
  \qquad
  s := \frac{da}{a} - t.
\end{equation*}
Then we have:
\begin{equation*}
  \bM_{j}
  =
  \begin{bmatrix}
    \bx\bbeta^\top & \bx \\
    d\bbeta^\top + \dfrac{\bbeta^\top}{a}da - t\bbeta^\top & s
  \end{bmatrix}.
\end{equation*}
For a partitioned matrix 
\begin{equation*}
\bM = \begin{bmatrix} \P & \q \\ \r^\top & s \end{bmatrix},
\end{equation*}
one has,
\begin{equation*}
  \mathrm{Tr}(\bM^2) = \mathrm{Tr}(\P^2) + 2\r^\top\q + s^2.
\end{equation*}
Here,
\begin{equation*}
  \P = \bx\bbeta^\top,
  \qquad \q=\bx,
  \qquad \r^\top = d\bbeta^\top + \frac{\bbeta^\top}{a}da - t\bbeta^\top.
\end{equation*}
Since $\P$ is rank one, we have:
\begin{equation*}
  \P^2 = \bx(\bbeta^\top \bx)\bbeta^\top = t\,\bx\bbeta^\top,
  \qquad
  \mathrm{Tr}(\P^2)=t^2.
\end{equation*}
Also,
\begin{align*}
  \r^\top\q
  &= \left(d\bbeta^\top + \frac{\bbeta^\top}{a}da - t\bbeta^\top\right)\bx \\
  &= d\bbeta^\top \bx + \frac{da}{a}\,\bbeta^\top \bx - t\,\bbeta^\top \bx \\
  &= a\,d\bbeta^\top \S d\bbeta + \frac{da}{a}t - t^2.
\end{align*}
Finally,
\begin{equation*}
  s^2 = \left(\frac{da}{a} - t\right)^2
  = \left(\frac{da}{a}\right)^2 - 2\frac{da}{a}t + t^2.
\end{equation*}
Putting these together,
\begin{align*}
  \mathrm{Tr}(\bM_{j}^2)
  &= t^2 + 2\left(a\,d\bbeta^\top \S d\bbeta + \frac{da}{a}t - t^2\right)
     + \left(\frac{da}{a}\right)^2 - 2\frac{da}{a}t + t^2 \\
  &= \left(\frac{da}{a}\right)^2 + 2a\,d\bbeta^\top \S d\bbeta.
\end{align*}
Since $ds^2_{\mathrm{FR}} = \frac12\mathrm{Tr}\{(\bSigma d\bTheta)^2\}$, the contribution from step $j$ is:
\begin{eqnarray}
  ds^2_{\mathrm{FR},j}
  &=& \frac12\left(\frac{d\tilde\theta_{jj}}{\tilde\theta_{jj}}\right)^2
  + \tilde\theta_{jj}\, d\bbeta_{\bullet j}^\top \bSigma_{j-1} \, d\bbeta_{\bullet j}.\label{eq:frj}
\end{eqnarray}
Let $s_{j}= \log \tilde\theta_{jj}$ and note that $\mathbf{r}_{j} = \tilde\theta_{jj}\bbeta_{\bullet j} = \exp(s_{j})\bbeta_{\bullet j}$. Then,
\begin{eqnarray}
d s_{j}&=& d \tilde\theta_{jj}/\tilde\theta_{jj} \label{eq:1}\\
d\bbeta_{\bullet j}&=&\exp(-s_{j})(d \mathbf{r}_{j} - \mathbf{r}_{j} ds_{j}). \nonumber
\end{eqnarray}
Thus,
\begin{align}
\tilde\theta_{jj}\, d\bbeta_{\bullet j}^\top \bSigma_{j-1} \, d\bbeta_{\bullet j} &=\exp(s_{j})\, d\bbeta_{\bullet j}^\top \bSigma_{j-1} \, d\bbeta_{\bullet j} \nonumber\\
&=\exp(-s_{j})(d \mathbf{r}_{j} - \mathbf{r}_{j} ds_{j})^\top\bSigma_{j-1} (d \mathbf{r}_{j} - \mathbf{r}_{j} ds_{j}).\label{eq:2}
\end{align}
Putting \eqref{eq:1} and ~\eqref{eq:2} into~\eqref{eq:frj} yields:
\begin{eqnarray}
  ds^2_{\mathrm{FR},j}
  &=&  \frac12 ds^2_{j}
  + \exp(-s_{j})\, (D \mathbf{r}_{j})^\top \bSigma_{j-1} \, D\mathbf{r}_{j},\label{eq:frjj}
\end{eqnarray}
where,
$$
D\mathbf{r}_{j}= d \mathbf{r}_{j} - \mathbf{r}_{j} ds_{j}.
$$
Summing over $j=2,\dots,p$, together with the initial scalar step $j=1$ giving $ds_1 = d \tilde\theta_{11}/\tilde\theta_{11}$, yields the result.

\section{Proof of Proposition~\ref{prop:frchol}}\label{supp:frchol}
Recall that:
\[
\bTheta_j=\bL_j \bL_j^\top,
\qquad
 d\bTheta_j=d\bL_j\,\bL_j^\top+\bL_j\,d\bL_j^\top.
\]
Hence,
\[
\bTheta_j^{-1}d\bTheta_j
=\bL_j^{-\top}\bL_j^{-1}(d\bL_j\, \bL_j^\top+L_j\,d\bL_j^\top)
=\bL_j^{-\top}(\A_j+\A_j^\top)\bL_j^\top,
\]
where $\A_j=\bL_j^{-1}d\bL_j$. By cyclicity of trace,
\[
\operatorname{tr}\bigl((\bTheta_j^{-1}d\bTheta_j)^2\bigr)
=\operatorname{tr}\bigl((\A_j+\A_j^\top)^2\bigr),
\]
and so,
\[
 ds_{FR,j}^2=\frac12\operatorname{tr}\bigl((\A_j+\A_j^\top)^2\bigr).
\]
We now compute $\A_j$ recursively. Since,
\[
\bL_j=
\begin{bmatrix}
\bL_{j-1} & 0\\
\bell_j^\top & \alpha_j
\end{bmatrix},
\]
its inverse is:
\[
\bL_j^{-1}=
\begin{bmatrix}
\bL_{j-1}^{-1} & 0\\
-\alpha_j^{-1}\bell_j^\top \bL_{j-1}^{-1} & \alpha_j^{-1}
\end{bmatrix}.
\]
Also,
\[
d\bL_j=
\begin{bmatrix}
d\bL_{j-1} & 0\\
d\bell_j^\top & d\alpha_j
\end{bmatrix}.
\]
Therefore,
\[
\A_j=\bL_j^{-1}d\bL_j
=
\begin{bmatrix}
\bL_{j-1}^{-1}d\bL_{j-1} & 0\\
-\alpha_j^{-1}\bell_j^\top \bL_{j-1}^{-1}d\bL_{j-1}+\alpha_j^{-1}d\bell_j^\top & \alpha_j^{-1}d\alpha_j
\end{bmatrix}.
\]
Writing:
\[
\A_{j-1}:=\bL_{j-1}^{-1}d\bL_{j-1},
\qquad
c_j:=\frac{d\alpha_j}{\alpha_j},
\qquad
\b_j^\top:=\alpha_j^{-1}(d\bell_j^\top-\bell_j^\top \bL_{j-1}^{-1}d\bL_{j-1}),
\]
we have the compact block form:
\[
\A_j=
\begin{bmatrix}
\A_{j-1} & 0\\
\b_j^\top & c_j
\end{bmatrix}.
\]
Hence,
\[
\A_j+\A_j^\top=
\begin{bmatrix}
\S_{j-1} & \b_j\\
\b_j^\top & 2c_j
\end{bmatrix},
\qquad
\S_{j-1}:=\A_{j-1}+\A_{j-1}^\top.
\]
Squaring this block matrix:
\[
\begin{bmatrix}
\S_{j-1} & \b_j\\
\b_j^\top & 2c_j
\end{bmatrix}^2
=
\begin{bmatrix}
\S_{j-1}^2+\b_j\b_j^\top & \S_{j-1}\b_j+2c_j \b_j\\
\b_j^\top \S_{j-1}+2c_j \b_j^\top & \b_j^\top \b_j+4c_j^2
\end{bmatrix}.
\]
Taking traces gives,
\[
\operatorname{tr}\bigl((\A_j+\A_j^\top)^2\bigr)
=\operatorname{tr}(\S_{j-1}^2)+2\b_j^\top \b_j+4c_j^2.
\]
Therefore,
\[
 ds_{FR,j}^2
 =\frac12\operatorname{tr}(\S_{j-1}^2)+\|\b_j\|^2+2c_j^2
 =ds_{FR,j-1}^2+\|\b_j\|^2+2\left(\frac{d\alpha_j}{\alpha_j}\right)^2.
\]
This proves the recursive formula.

\section{Proof of Proposition~\ref{prop:frv}}\label{sec:frv}
For $j\ge 2$, recall from~\eqref{eq:G} that the Fisher--Rao metric block corresponding to the slice $(s_j,\r_j)$ is:
\[
\G_{\mathrm{FR}} (s_j,\r_j)=
\begin{bmatrix}
\frac12+e^{-s_{j}}\r_{j}^\top\bSigma_{j-1}r_{j}
&
-e^{-s_{j}}\r_{j}^\top\bSigma_{j-1}
\\[1mm]
-e^{-s_{j}}\bSigma_{j-1}\r_{j}
&
e^{-s_{j}}\bSigma_{j-1}
\end{bmatrix},
\]
For $j=1$, the metric contributes only the scalar block $1/2$. Since the full Fisher--Rao metric is block-diagonal by slices in the coordinates $(\s,\r)$, its determinant is the product of the determinants of these blocks. 

Fix $j\ge 2$. Write,
\[
\A:=\frac12+e^{-s_j}r_j^\top\bSigma_{j-1}\r_j,
\qquad
\b:=-e^{-s_j}\bSigma_{j-1}\r_j,
\qquad
\C:=e^{-s_j}\Sigma_{j-1}.
\]
Then,
\[
\G_{\mathrm{FR}} (s_j,\r_j)=
\begin{bmatrix}
\A & \b^\top\\
\b & \C
\end{bmatrix}.
\]
By the Schur determinant formula,
\[
\det(\G_{\mathrm{FR}} (s_j,\r_j))=\det(\C)\bigl(\A-\b^\top \C^{-1}\b\bigr).
\]
Now,
\[
\C^{-1}=e^{s_j}\bSigma_{j-1}^{-1}=e^{s_j}\bTheta_{j-1}.
\]
Thus,
\[
\b^\top \C^{-1}\b
=
\bigl(-e^{-s_j}\r_j^\top\bSigma_{j-1}\bigr)
\bigl(e^{s_j}\bTheta_{j-1}\bigr)
\bigl(-e^{-s_j}\bSigma_{j-1}\r_j\bigr)
=e^{-s_j}\r_j^\top\bSigma_{j-1}r_j.
\]
Hence,
\[
\A-\b^\top \C^{-1}\b
=
\frac12+e^{-s_j}\r_j^\top\bSigma_{j-1}r_j-e^{-s_j}\r_j^\top\bSigma_{j-1}\r_j
=
\frac12,
\]
and therefore,
\[
\det(\G_{\mathrm{FR}} (s_j,\r_j))
=
\frac12\det(\C)
=
\frac12 e^{-(j-1)s_j}\det(\bSigma_{j-1}),
\]
since $\bSigma_{j-1}$ is a $(j-1)\times (j-1)$ matrix. Including the scalar block $1/2$ from $s_1$, we get,
\[
\det(\G_{\mathrm{FR}} (\s,\r))
=
\frac12\prod_{j=2}^p\det(\G_{\mathrm{FR}} (s_j,\r_j))
=
2^{-p}\prod_{j=2}^p e^{-(j-1)s_j}\det(\bSigma_{j-1}).
\]
Since $\bSigma_{j-1}=\bTheta_{j-1}^{-1}$,
\[
\det(\bSigma_{j-1})=\det(\bTheta_{j-1})^{-1}
=
\exp\!\left(-\sum_{k=1}^{j-1}s_k\right).
\]
Substituting this yields,
\[
\det(\G_{\mathrm{FR}} (\s,\r))
=
2^{-p}
\exp\!\left(-\sum_{j=2}^p\Bigl((j-1)s_j+\sum_{k=1}^{j-1}s_k\Bigr)\right).
\]
It remains to simplify the exponent. Observe that
\[
\sum_{j=2}^p \sum_{k=1}^{j-1}s_k
=
\sum_{k=1}^{p-1}(p-k)s_k,
\]
so the total exponent is,
\[
\sum_{j=2}^p (j-1)s_j + \sum_{k=1}^{p-1}(p-k)s_k
=
\sum_{k=1}^{p-1}(p-1)s_k + (p-1)s_p
=
(p-1)\sum_{k=1}^p s_k.
\]
Thus,
\[
\det(\G_{\mathrm{FR}} (\s,\r))=2^{-p}e^{-(p-1)v},
\qquad
v:=\sum_{k=1}^p s_k.
\]
Taking square roots gives,
\[
d\mathrm{Vol}_{FR}=2^{-p/2}e^{-(p-1)v/2}\,ds\,dr.
\]
Since $s_j=d_j + v/p$, the transformation $(v,\d)\leftrightarrow \s$ is linear with determinant $1$. Thus, the same formula holds in $(v,\bd,\r)$ coordinates:
\[
d\mathrm{Vol}_{FR}=2^{-p/2}e^{-(p-1)v/2}\,dv\,d\bd\,d\r.
\]
Finally,  the Jacobian formula, $J(\bx\to\bTheta)=\exp(v)$, yields:
\[
d\bTheta=e^v\,dv\,d\bd\,d\r.
\]
Substituting this into the preceding display gives,
\[
d\mathrm{Vol}_{FR}(\bTheta)
=
2^{-p/2}e^{-(p-1)v/2}e^{-v}\,d\bTheta
=
2^{-p/2}e^{-(p+1)v/2}\,d\bTheta.
\]
Since $e^v=\det(\bTheta)$, this becomes,
\[
d\mathrm{Vol}_{FR}(\bTheta)
=
2^{-p/2}(\det\bTheta)^{-(p+1)/2}\,d\bTheta,
\]
as claimed.

\section{Proof of Proposition~\ref{prop:laplace}}\label{sec:laplace}
We prove each of the three displayed equations.\\
\\
1. (Derivation of the line element in~\eqref{eq:metric-vdbeta}). Recall,
$\r_j=e^{s_j}\bbeta_j$ where $s_j=d_j + v/p$. Differentiating,
\[
d\r_j=e^{s_j}(d\bbeta_j+\bbeta_j\,ds_j),
\]
and hence
\[
D\r_j
=
d\r_j-\r_j\,ds_j
=
e^{s_j}(d\bbeta_j+\bbeta_j\,ds_j)-e^{s_j}\bbeta_j\,ds_j
=
e^{s_j}d\bbeta_j.
\]
Substituting into \eqref{eq:fr} gives,
\begin{equation}
\label{eq:metric-sbeta}
ds_{FR}^2
=
\frac12\sum_{j=1}^p (ds_j)^2
+
\sum_{j=2}^p e^{s_j}(d\bbeta_j)^\top \bSigma_{j-1}d\bbeta_j.
\end{equation}
Observe that:
\[
ds_j=dd_j+\frac{dv}{p}.
\]
Since \(\sum_{j=1}^p dd_j=0\), we obtain
\[
\sum_{j=1}^p (ds_j)^2
=
\sum_{j=1}^p \Bigl(dd_j+\frac{dv}{p}\Bigr)^2
=
\sum_{j=1}^p (dd_j)^2
+
\frac{2dv}{p}\sum_{j=1}^p dd_j
+
\sum_{j=1}^p \frac{dv^2}{p^2}
=
\sum_{j=1}^p (dd_j)^2+\frac{dv^2}{p}.
\]
Hence \eqref{eq:metric-sbeta} becomes
\[
ds_{FR}^2
=
\frac1{2p}dv^2
+
\frac12\sum_{j=1}^p (dd_j)^2
+
\sum_{j=2}^p e^{d_j+v/p}(d\bbeta_j)^\top \bSigma_{j-1}d\bbeta_j,
\]
which is \eqref{eq:metric-vdbeta}.\\

\noindent 2. (Derivation of the metric tensor in $\G^{-1}$ in \eqref{eq:inverse-metric-vdbeta}). It is clear that the metric given by the line element in \eqref{eq:metric-vdbeta} is block diagonal. Now, recall,
\[
\mathbf d=(d_2,\dots,d_p)^\top,\qquad d_1=-\sum_{j=2}^p d_j.
\]
Thus,
\[
\sum_{j=1}^p (dd_j)^2
=
\Bigl(\sum_{j=2}^p dd_j\Bigr)^2+\sum_{j=2}^p (dd_j)^2
=
d\mathbf d^\top (I_{p-1}+\mathbf 1\mathbf 1^\top)\, d\mathbf d.
\]
Therefore, in matrix form,
\[
ds_{FR}^2
=
\frac1{2p}dv^2
+
\frac12\,d\mathbf d^\top (I_{p-1}+\mathbf 1\mathbf 1^\top)\,d\mathbf d
+
\sum_{j=2}^p e^{d_j+v/p}(d\beta_j)^\top \Sigma_{j-1}d\beta_j.
\]
We proceed to identify the blocks of the metric tensor $\G$ and the respective inverses.\\

\noindent (i) The \(v\)-block is \(\frac1{2p}\), whose inverse is \(2p\).\\

\noindent (ii) For the \(\bd\)-block, set
\[
\A:=I_{p-1}+\mathbf 1\mathbf 1^\top.
\]
A direct computation shows,
\[
\A^{-1}=I_{p-1}-\frac1p \mathbf 1\mathbf 1^\top=:\P.
\]
Because the \(d\)-metric block is \(\frac12 \A\), its inverse is \(2\P\).\\

\noindent (iii) Finally, for each \(j\ge2\), the \(\bbeta_j\)-block of the metric is,
\[
e^{d_j+v/p}\bSigma_{j-1},
\]
whose inverse is
\[
e^{-(d_j+v/p)}\bTheta_{j-1}.
\]
This proves \eqref{eq:inverse-metric-vdbeta}.\\

\noindent 3. (Derivation of the Laplace--Beltrami operator in \eqref{eq:LB-expanded-vdbeta}). Recall from Proposition~\ref{prop:frv},
\[
d\Vol_{FR}
=
2^{-p/2}e^{-(p-1)v/2}\,dv\,d\bd\,d\r.
\]
We now compute the Jacobian of the transformation \(\r\mapsto \bbeta\). Since \(\r_j=e^{s_j}\bbeta_j\) and \(\bbeta_j\in\mathbb R^{j-1}\),
\[
d\r_j = e^{(j-1)s_j}\,d\bbeta_j
\]
at the Jacobian level. Therefore
\[
d\r
=
\prod_{j=2}^p e^{(j-1)s_j}\,d\bbeta
=
\exp\!\Bigl(\sum_{j=2}^p (j-1)s_j\Bigr)\,d\bbeta.
\]
Using \(s_j=d_j+\frac{v}{p}\), and substituting into the volume form,
\[
d\Vol_{FR}
=
2^{-p/2}
e^{-(p-1)v/2}
\exp\!\Bigl(\sum_{j=2}^p (j-1)d_j+\frac{(p-1)v}{2}\Bigr)
\,dv\,d\bd\,d\bbeta.
\]
The \(v\)-factors cancel exactly, leaving,
\[
d\Vol_{FR}
=
2^{-p/2}\exp\!\Bigl(\sum_{j=2}^p (j-1)d_j\Bigr)\,dv\,d\bd\,d\bbeta.
\]
For a Riemannian metric \(\G\), recall that the Laplace--Beltrami operator is:
\[
\Delta_{FR}f
=
\frac1{\sqrt{\det \G}}
\partial_\alpha\Bigl(\sqrt{\det \G}\,\G^{\alpha\beta}\partial_\beta f\Bigr).
\]
Let $\mu:=\mu(\bd)= 2^{-p/2}\exp\!\Bigl(\sum_{j=2}^p (j-1)d_j\Bigr)$. Using the block diagonal inverse metric \eqref{eq:inverse-metric-vdbeta}, we obtain,
\begin{align}
\Delta_{FR} f
&=
\frac1\mu \partial_v\bigl(\mu\,2p\,\partial_v f\bigr)
+
\frac1\mu \nabla_{\mathbf d}\cdot\bigl(\mu\,2\P\,\nabla_{\mathbf d}f\bigr)
+
\sum_{j=2}^p
\frac1\mu \nabla_{\bbeta_j}\cdot\Bigl(\mu\,e^{-(d_j+v/p)}\bTheta_{j-1}\nabla_{\bbeta_j}f\Bigr),\label{eq:hess}
\end{align}
Since \(\mu\) depends only on \(\mathbf d\), we have
\begin{align}
\frac1\mu \partial_v(\mu\,2p\,\partial_v f)&=2p\,\partial_{vv}f.\label{eq:hessv}
\end{align}
Also,
\[
\nabla_{\mathbf d}\log\mu
=
(1,\ldots,p-1)^\top.
\]
Therefore,
\[
\frac1\mu \nabla_{\mathbf d}\cdot(\mu\,2\P\nabla_{\mathbf d}f)
=
2\,\nabla_{\mathbf d}\cdot(P\nabla_{\mathbf d}f)
+
2(P\nabla_{\bd}\log\mu)^\top \nabla_{\mathbf d}f.
\]
Since $\nabla_\bd \log \mu = (1,\ldots, p-1)^\top$, we have that $\P\nabla_\bd \log \mu$ has $j$th component given by $\sum_{k=1}^{p-1}(\delta_{jk} - 1/p)(k-1) = (j-1)- (1/p)\sum_{k=1}^{p-1} (k-1) = (j-1) - (p-1)/2$. Thus, the \(j\)-th component of \(2P\nabla_{\bd}\log\mu\) is
$
(2j-p-1)$ for $j=2,\ldots,p$. 
Hence,
\begin{align}
\frac1\mu \nabla_{\mathbf d}\cdot(\mu\,2P\nabla_{\mathbf d}f)
=
2\sum_{i,j=2}^p\Bigl(\delta_{ij}-\frac1p\Bigr)\partial_{d_i d_j}f
+
\sum_{j=2}^p (2j-p-1)\partial_{d_j}f.\label{eq:hessd}
\end{align}
Finally, since \(\mu\) and \(\bTheta_{j-1}\) are independent of \(\bbeta_j\),
\begin{align}
\frac1\mu \nabla_{\bbeta_j}\cdot\Bigl(\mu\,e^{-(d_j+v/p)}\bTheta_{j-1}\nabla_{\bbeta_j}f\Bigr)
&=
e^{-(d_j+v/p)}\tr\!\bigl(\bTheta_{j-1}H_{\bbeta_j}f\bigr).\label{eq:hessbeta}
\end{align}
Combining ~\eqref{eq:hessv}, \eqref{eq:hessd}, \eqref{eq:hessbeta} into \eqref{eq:hess}, yields the desired expression \eqref{eq:LB-expanded-vdbeta}.

\section{Proof of Corollary \ref{cor:laplace-v-d-beta}}\label{sec:action}
We apply the Laplace--Beltrami operator from~\eqref{eq:LB-expanded-vdbeta} successively to $(v,\bd,\bbeta)$.\\

\noindent (i) For \(f=v\), all derivatives vanish except \(\partial_v f=1\), but \(\partial_{vv}f=0\). Since \eqref{eq:LB-expanded-vdbeta} contains no first-order \(v\)-drift term, we obtain
\[
\Delta_{FR}v=0.
\]

\noindent (ii) Next, let \(f=d_j\) for some fixed \(j\in\{2,\dots,p\}\). Then
\[
\partial_{d_j}f=1,\qquad \partial_{d_i d_j}f=0,\qquad
\partial_v f=0,\qquad H_{\bbeta_k}f=0 \ \text{ for all }k.
\]
Substituting into \eqref{eq:LB-expanded-vdbeta} yields
\[
\Delta_{FR}d_j=2j-p-1.
\]

\noindent (iii) Finally, let \(f=\bbeta_{j,a}\), the \(a\)-th component of \(\bbeta_j\). Then \(f\) is linear in \(\beta_j\), independent of \(v\), and independent of all \(d_k\). Hence
\[
\partial_v f=0,\qquad \partial_{d_k}f=0,\qquad
H_{\bbeta_k}f=0 \ \text{ for all }k,
\]
because \(H_{\bbeta_j}f=0\) for a linear function and \(H_{\bbeta_k}f=0\) trivially for \(k\neq j\). Therefore every term in \eqref{eq:LB-expanded-vdbeta} vanishes, and
\[
\Delta_{FR}\bbeta_{j,a}=0.
\]
The proof of the corollary is complete.

\section{Proof of Lemma~\ref{prop:diff}}\label{sec:diff}
For an It\^o SDE:
\[
d\bX_t=\b(\bX_t)\,dt+\bsigma(\bX_t)\,d\W_t,
\]
the generator is,
\[
Lf = \b\cdot \nabla f + \frac12(\bsigma\bsigma^\top)^{\alpha\beta}\partial_{\alpha\beta}f.
\]
Therefore, by \eqref{eq:inverse-metric-vdbeta}, to realize an intrinsic Brownian motion with \(L_{\mathrm{BM}}=\frac12\Delta_{FR}\), it suffices to choose \(\bsigma\) so that,
\[
\bsigma\bsigma^\top
=
\diag\!\Bigl(
2p,\;
2\P,\;
e^{-(d_2+v/p)}\bTheta_1,\;
\dots,\;
e^{-(d_p+v/p)}\bTheta_{p-1}
\Bigr),
\]
together with drift \(\frac12\boldsymbol{\kappa}\) in the \(\mathbf d\)-block and zero drift elsewhere, by Corollary~\ref{cor:laplace-v-d-beta}. This yields \eqref{eq:gen-v}--\eqref{eq:gen-beta}.

For the more general diffusion with prescribed drift field \(\B\), we simply add \(\B\) to the Brownian drift. The resulting generator is
\[
Lf=\B\cdot\nabla f+\frac12\Delta_{FR}f,
\]
which proves \eqref{eq:gen-generator}.

Finally, the intrinsic Langevin diffusion associated with target density \(\pi\) relative to the Riemannian volume is the diffusion with generator
\[
L_{\mathrm{Lan}}f
=
\frac12\Delta_{FR}f+\frac12\langle \nabla_{FR}\log\pi,\nabla_{FR}f\rangle_{FR}.
\]
In coordinates, \(\nabla_{FR}\log\pi = \G_{FR}^{-1}\nabla \log\pi\). Hence,
\[
\B=\frac12\,\G_{FR}^{-1}\nabla\log\pi.
\]
Using the inverse metric from Proposition~\ref{prop:laplace}, we obtain:
\[
B_v = p\,\partial_v\log\pi,\qquad
B_\bd = P\nabla_{\mathbf d}\log\pi,\qquad
B_{\bbeta_j}=\frac12 e^{-(d_j+v/p)}\bTheta_{j-1}\nabla_{\bbeta_j}\log\pi.
\]
Substituting these into \eqref{eq:gen-v}--\eqref{eq:gen-beta} yields \eqref{eq:langevin-Bv}--\eqref{eq:langevin-Bbeta}.

\section{Proof of Proposition~\ref{prop:ham}}\label{sec:ham}
Write $\bx_{\mathrm{norm}}=(v,\bxi)$ with $\bxi=(\bd,\r_{\mathrm{norm}})$. Since by the continuity equation,
\[
\partial_t q_t = -\nabla_{\bx_{\mathrm{norm}}}\cdot(q_t \u),
\]
the standard identity for entropy relative to a reference density $m(\bx_{\mathrm{norm}})$,
\[
H_m(q_t):=-\int q_t(\bx_{\mathrm{norm}})\log\!\left(\frac{q_t(\bx_{\mathrm{norm}})}{m(\bx_{\mathrm{norm}})}\right)d{\bx_{\mathrm{norm}}},
\]
yields:
\begin{align}
\frac{d}{dt}H_m(q_t)
&= \mathbb E_{q_t}[\nabla_{\bx_{\mathrm{norm}}}\cdot \u] + \mathbb E_{q_t}[\u\cdot \nabla_{\bx_{\mathrm{norm}}}\log m],\label{eq:ent}
\end{align}
provided boundary terms vanish under integration by parts. We further have:
\begin{align}
\nabla_{\bx_{\mathrm{norm}}}\cdot \u &= \partial_v u(v,t) + \nabla_{\bxi}\cdot \u(\bxi,t; v) = \partial_v u(v,t),\label{eq:ham1}
\end{align}
since by the Hamiltonian assumption $\nabla_{\bxi}\cdot \u(\bxi,t; v)\equiv 0$. By the RT Jacobian formula, for the ambient SPD volume, the reference density is:
\[
m_{\bTheta}(\bx_{\mathrm{norm}})=\exp((p+1)v/2),
\qquad \nabla_{\bx_{\mathrm{norm}}}\log m_{\bTheta}=((p+1)/2,0,\ldots,0).
\]
Thus,
\begin{align}
\u\cdot \nabla_\bx\log m_{\bTheta}&=\frac{p+1}{2}  u(v,t).\label{eq:ham2}
\end{align}
Using~\eqref{eq:ham1} and~\eqref{eq:ham2} into~\eqref{eq:ent},
\[
\frac{d}{dt}H_{\bTheta}(q_t)=\mathbb E_{q_t}[\partial_v u(v,t)+ \frac{p+1}{2} u(v,t)],
\]
which proves Part~(i).

For the Fisher--Rao volume, the reference density is:
\[
m_{FR}(\bx_{\mathrm{norm}})=2^{-p/2},
\qquad \nabla_{\bx_{\mathrm{norm}}}\log m_{FR}=\left(0,\ldots,0\right).
\]
Therefore,
\[
\u\cdot \nabla_{\bx_{\mathrm{norm}}}\log m_{FR}=0,
\]
and so,
\[
\frac{d}{dt}H_{FR}(q_t)=\mathbb E_{q_t}\!\left[\partial_v u(v,t)\right].
\]
This proves Part~(ii).

\section{Proof of Proposition \ref{prop:ellipsoid-bounds}}\label{app:ellipsoid-bounds}
Recall the notation:
\[
s_j=\log \tilde\theta_{jj}=d_j+\frac{v}{p},\qquad \bbeta_j=\frac{\tilde\theta_{\bullet j}}{\tilde\theta_{jj}},
\]
and recall from \eqref{eq:metric-vdbeta} that the Fisher--Rao line element in $(v,\bd,\bbeta)$ is:
\[
ds^2_{FR}
=
\frac{1}{2p}dv^2
+
\frac12\sum_{j=1}^p (dd_j)^2
+
\sum_{j=2}^p
e^{s_j}(d\bbeta_j)^\top \bSigma_{j-1}d\bbeta_j .
\]
Thus it remains only to bound the last term. Recall from \eqref{eq:recursion} that at stage $j$ one has:
\[
\bTheta_j
=
\begin{bmatrix}
\bTheta_{j-1}+\tilde\theta_{jj}\bbeta_j\bbeta_j^\top & \tilde\theta_{jj}\bbeta_j\\
\tilde\theta_{jj}\bbeta_j^\top & \tilde\theta_{jj}
\end{bmatrix},
\qquad
\bTheta_1=(\tilde\theta_{11}),
\]
and  \(\bSigma_{j-1}=\bTheta_{j-1}^{-1}\). Since \(\bTheta_j\) is obtained recursively from \(\bTheta\) by rank-one Schur complements, the spectral bound:
\[
mI_p\preceq \bTheta \preceq MI_p,
\]
implies, recursively, that:
\[
mI_j\preceq \bTheta_j\preceq MI_j,\qquad j=1,\ldots,p.
\]
To see this, indeed, if,
\[
A=\begin{bmatrix}
A_{11} & A_{12}\\
A_{21} & A_{22}
\end{bmatrix},
\]
satisfies \(mI\preceq A\preceq MI\), then the Schur complement:
\[
A/A_{22}=A_{11}-A_{12}A_{22}^{-1}A_{21},
\]
satisfies the same bounds. The lower bound follows from the variational identity:
\[
x^\top(A/A_{22})x
=
\min_z
\begin{pmatrix}x\\z\end{pmatrix}^{\!\top}
A
\begin{pmatrix}x\\z\end{pmatrix}
\ge m\|x\|^2,
\]
while the upper bound follows from:
\[
A/A_{22}\preceq A_{11}\preceq MI.
\]
Applying this argument repeatedly along the reverse telescoping recursion gives the displayed
bounds for all \(\bTheta_j\). Consequently,
\[
mI_{j-1}\preceq \bTheta_{j-1}\preceq MI_{j-1},
\quad\quad
\frac1M I_{j-1}\preceq \bSigma_{j-1}=\bTheta_{j-1}^{-1}
\preceq \frac1m I_{j-1}.
\]
Moreover, since \(\tilde\theta_{jj}\) is itself the final Schur pivot at stage \(j\), the same Schur-residual bounds give:
\[
m\le \tilde\theta_{jj}\le M .
\]
Therefore, for any increment \(\u_j=d\bbeta_j\),
\[
\frac{m}{M}\|\u_j\|^2
\le
e^{s_j}\u_j^\top\bSigma_{j-1}\u_j
\le
\frac{M}{m}\|\u_j\|^2 .
\]
Substituting \(\u_j=d\bbeta_j\) into the Fisher--Rao line element yields:
\[
\frac{1}{2p}dv^2
+
\frac12\sum_{j=1}^p (dd_j)^2
+
\frac{m}{M}\sum_{j=2}^p \|d\bbeta_j\|^2
\le
ds^2_{FR}
\le
\frac{1}{2p}dv^2
+
\frac12\sum_{j=1}^p (dd_j)^2
+
\frac{M}{m}\sum_{j=2}^p \|d\bbeta_j\|^2,
\]
which proves \eqref{eq:metric-sandwich}.

\end{document}

%% file: math_commands_JMLR.tex
\newcommand{\tr}{{\rm tr}}

\renewcommand{\a}{{\bf a}}
\renewcommand{\b}{{\bf b}}

\renewcommand{\d}{{\rm d}}  % for derivatives
\renewcommand{\r}{{\bf r}}
\newcommand{\s}{{\bf s}}

\renewcommand{\u}{{\bf u}}

\renewcommand{\H}{{\bf H}}
\renewcommand{\L}{{\bf L}}

\renewcommand{\P}{{\bf P}}
\newcommand{\Q}{{\bf Q}}

\renewcommand{\S}{{\bf S}}
\newcommand{\T}{{\bf T}}

\newcommand{\W}{{\bf W}}
\newcommand{\bsigma}{\boldsymbol{\sigma}}

\newcommand{\bbeta}{\boldsymbol{\beta}}

\newcommand{\bLambda}{\mathbf{\Lambda}}

\newcommand{\btheta}{\boldsymbol{\theta}}

\newcommand{\bSigma}{\boldsymbol{\Sigma}}

\newcommand{\bomega}{\boldsymbol{\omega}}
\newcommand{\ben}{\begin{enumerate}}
\newcommand{\een}{\end{enumerate}}
\newcommand{\beq}{\begin{equation}}
\newcommand{\eeq}{\end{equation}}
\newcommand{\bde}{\begin{description}}
\newcommand{\ede}{\end{description}}

\newcommand{\bx}{{\bf x}}

\newcommand{\bd}{{\bf d}}
\newcommand{\bX}{{\bf X}}
\newcommand{\by}{{\bf y}}

\usepackage{booktabs,array}

\newcount\rowc